\documentclass[lettersize,journal]{IEEEtran}
\usepackage{amsmath,amsfonts}
\usepackage{algorithmic}
\usepackage{algorithm}
\usepackage{array}
\usepackage[caption=false,font=normalsize,labelfont=sf,textfont=sf]{subfig}
\usepackage{textcomp}
\usepackage{stfloats}
\usepackage{url}
\usepackage{verbatim}
\usepackage{graphicx}
\usepackage{cite}
\usepackage{multirow}
\usepackage{booktabs}
\usepackage{enumitem}
\usepackage{xcolor}
\newcommand{\gainpos}[1]{\textcolor{red}{\scriptsize(#1)}}
\newcommand{\gainneg}[1]{\textcolor{green!50!black}{\scriptsize(#1)}}

\begin{document}

\title{Dynamic Interpolation for Training-Time Prediction Reshaping in Low-Light Image Enhancement}

\author{Peirui Zhu, Shijie Hao, Jingxi Liao, Richang Hong,~\IEEEmembership{Senior Member,~IEEE}, and Meng Wang,~\IEEEmembership{Fellow,~IEEE}
\thanks{Peirui Zhu, Shijie Hao, Jingxi Liao, Richang Hong and Meng Wang are with the School of Computer Science and Information Engineering, Hefei University of Technology, Hefei 230009, China. Shijie Hao is the corresponding author (hfut.hsj@gmail.com).}}


\IEEEpubid{
\begin{minipage}{\textwidth}
\centering
\scriptsize
Copyright \copyright\ 2026 IEEE. Personal use of this material is permitted. However, permission to use this material for any other purposes must\\
be obtained from the IEEE by sending an email to pubs-permissions@ieee.org.
\end{minipage}
}

\maketitle

\begin{abstract}
Low-light image enhancement (LLIE) is an important image/video processing problem for reliable visual systems under poor illumination, yet it remains challenging due to the strong coupling among insufficient illumination, noise amplification, and structural degradation. Existing methods typically improve enhancement performance through either increasingly sophisticated network architectures or carefully engineered objective functions. While effective, the former often increases model complexity and deployment burden, whereas the latter usually relies on empirically tuned loss weights. In this paper, we revisit supervised LLIE from a training-dynamics perspective and observe that brightness-related errors tend to converge early, while residual noise and structural artifacts often remain insufficiently optimized in later stages. To address this issue, \textcolor{black}{we propose a Dynamic Interpolation Training-only Module (DITM), a lightweight training-only module that introduces no additional learnable parameters and adaptively reshapes the model prediction for supervised feed-forward LLIE training.} Specifically, DITM employs a brightness-aware interpolation scheme, in which the model prediction is combined with a mean brightness-aligned image constructed by rescaling the model prediction according to the mean brightness of the ground-truth image. By reducing the relative dominance of global brightness discrepancy, DITM makes residual local errors more influential during backpropagation after global luminance has largely stabilized, thereby facilitating artifact suppression and fine-detail restoration in later training stages. DITM is used only during training and is fully removed at inference, introducing no additional inference cost. Extensive experiments on multiple benchmark datasets and CNN-, Transformer-, and Mamba-based LLIE backbones show that DITM consistently improves fidelity and perceptual quality while preserving the inference-time architecture and introducing no additional inference-time computation.
\end{abstract}


\section{Introduction}
\IEEEPARstart {L}{ow-light} image enhancement (LLIE) is a fundamental image/video processing task for robust visual perception in poorly illuminated environments~\cite{zhao2025deep}. Unlike single-degradation problems such as denoising or super-resolution, LLIE must handle complex, coupled degradations \cite{li2021lowlight_survey,liu2021benchmarking}. In low-light environments, images typically exhibit both low dynamic range and elevated noise levels due to the low signal‑to‑noise ratio, which makes enhancement particularly challenging~\cite{su2022jnf}. This commonly leads to an \textit{“amplification-suppression dilemma”}: increasing illumination to restore visibility amplifies latent noise, whereas aggressive denoising risks erasing high-frequency structural details~\cite{fu2016lime,guo2017lime}. The main challenge lies in balancing global luminance restoration, local noise suppression, and structural fidelity under practical visual-system constraints.
\IEEEpubidadjcol
Existing LLIE methods often improve enhancement performance through network architecture design or specialized objective design. Network design typically focuses on enhancing the model's ability to handle complex degradations and capture global contextual relationships \cite{feijoo2025darkir,xu2022snr,pan2024illumination}. While effective, these methods may introduce additional computational overhead and inference latency, which is undesirable for resource-constrained visual processing systems. Another direction employs specialized loss terms to balance brightness enhancement, noise suppression, and detail preservation \cite{jiang2021enlightengan,ma2022toward,guo2020zero,lv2021attention}. Although effective, these methods remain highly sensitive to loss weight configuration, requiring extensive empirical tuning.

\begin{figure*}[t]
    \centering
    \includegraphics[width=\linewidth]{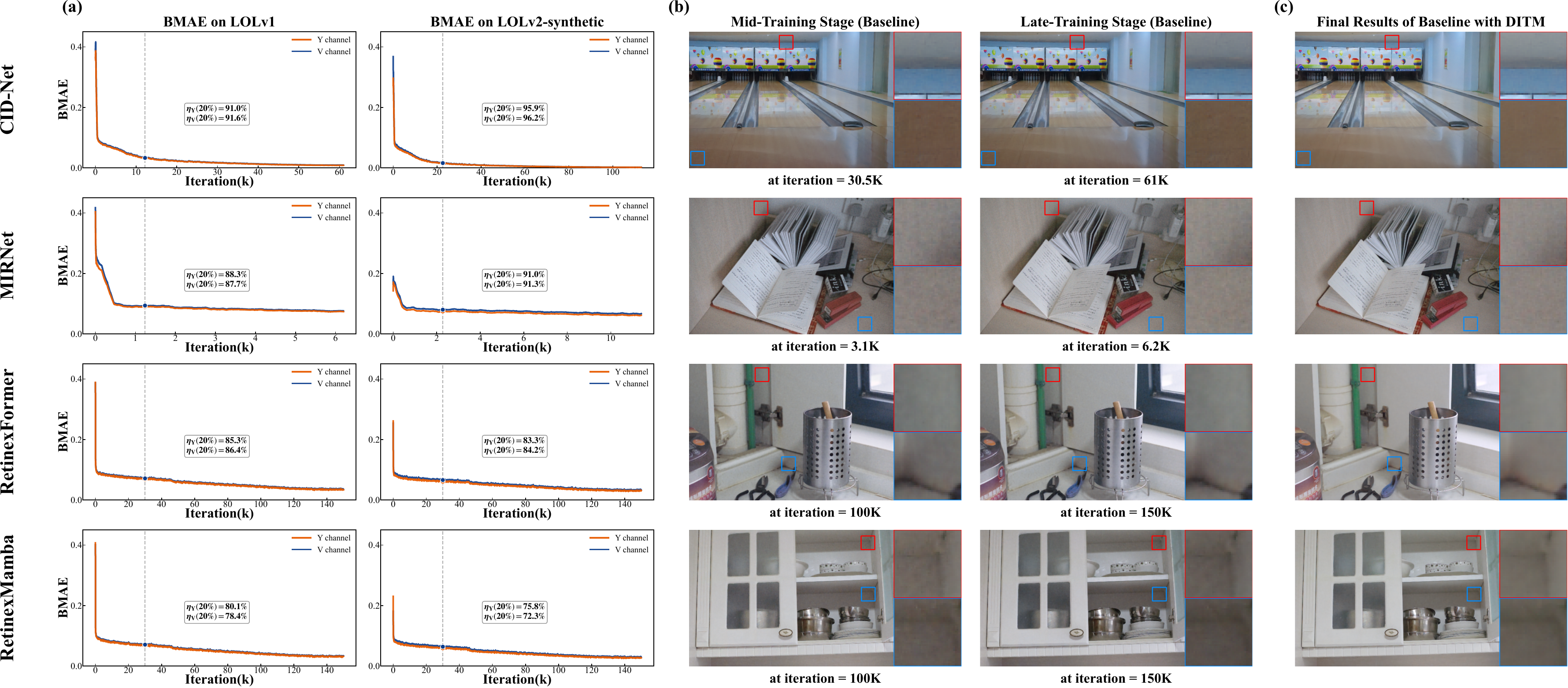}
    \caption{{\color{black}{
    Empirical motivation for DITM.
    \textbf{(a)} Global brightness discrepancy is reduced primarily during the early stage of baseline training. We present the evolution of the global brightness discrepancy throughout the training of four representative baseline LLIE models on LOLv1 and LOLv2-synthetic, evaluated using the Y channel of YCbCr and the V channel of HSV. Let $A_t$ denote the baseline prediction at training stage $t$, and let $Y$ denote the corresponding ground-truth image. For brightness representation $c$, BMAE is defined as
    $\mathrm{BMAE}_{c}(t)
    =
    \left|
    \mathrm{Mean}\!\left[\Phi_{c}(Y)\right]
    -
    \mathrm{Mean}\!\left[\Phi_{c}(A_t)\right]
    \right|$,
    where $\Phi_c(\cdot)$ extracts the specified brightness component,
    $\mathrm{Mean}[\cdot]$ computes its spatial mean over all pixels, and
    $c\in\{\mathrm{YCbCr\text{-}Y},\mathrm{HSV\text{-}V}\}$. At each training iteration, BMAE is computed by averaging the brightness discrepancy over the samples used for the current update.
    The early brightness-error reduction ratio is defined as
    $\eta_{c}(r)
    =
    \frac{
    \mathrm{BMAE}_{\mathrm{init}}^{c}
    -
    \mathrm{BMAE}_{r}^{c}
    }{
    \mathrm{BMAE}_{\mathrm{init}}^{c}
    -
    \mathrm{BMAE}_{\mathrm{final}}^{c}
    }
    \times 100\%$,
    where the three BMAE terms denote the averaged BMAE values around the initial stage, training progress $r$, and the final stage, respectively. The annotated $\eta_c(20\%)$ values show that 72.3\%--96.2\% of the overall BMAE reduction is already achieved within the first 20\% of training.
    \textbf{(b)} The baseline models still struggle to recover local structures and fine details as training approaches its final stage. The middle-stage baseline predictions contain noticeable residual noise, blurred structures, and local artifacts, some of which persist into the late stage. Motivated by these observations, we propose reshaping training-time predictions so that, after global brightness has largely stabilized, residual local restoration errors become relatively more influential during subsequent optimization, thereby facilitating late-stage artifact suppression and structural-detail restoration. \textbf{(c)} Visual examples of our DITM-based models. With DITM, the same backbones produce cleaner local structures and textures with fewer residual artifacts. In this figure, red and blue boxes indicate enlarged regions. Best viewed enlarged.}}}
    \label{fig:introduction}
\end{figure*}

{\color{black}

Although recent advances in network architectures and objective functions have substantially improved low-light image enhancement, we observe a recurring asynchronous convergence pattern during training. As shown in Fig.~\ref{fig:introduction}, the global brightness discrepancy is reduced mainly in the early stage, whereas residual noise, blurred structures, and local artifacts may persist until considerably later stages. To quantitatively characterize this optimization behavior, we track the evolution of the global brightness discrepancy throughout training across four representative LLIE models on LOLv1 and LOLv2-synthetic using two brightness representations. We evaluate this evolution using BMAE and the early reduction ratio $\eta_c(r)$. Across the evaluated settings, 72.3\%--96.2\% of the total BMAE reduction is achieved within the first 20\% of training (Fig.~\ref{fig:introduction}(a)), while noticeable local restoration errors may persist in late-stage predictions (Fig.~\ref{fig:introduction}(b)). These observations indicate an asynchronous optimization pattern: the image-level brightness discrepancy is largely reduced early, whereas local artifact suppression and structural-detail restoration continue to require further optimization. To address this issue, we propose reshaping training-time predictions to promote further optimization of the remaining restoration errors.

}

We propose the Dynamic Interpolation Training-only Module (DITM), a lightweight training-only module \textcolor{black}{that introduces no additional learnable parameters} for supervised feed-forward LLIE training.
Guided by an adaptive weight (interpolation factor), DITM generates a dynamically interpolated prediction by fusing the model prediction with a mean brightness-aligned image, which is constructed by rescaling the model prediction according to the mean brightness of the ground-truth image. This factor progressively adjusts the training signal by reducing the relative dominance of global luminance discrepancy in the reconstruction loss. As a result, after global brightness has largely stabilized, residual local errors receive relatively stronger influence during backpropagation, which helps the model further suppress artifacts and restore details. Importantly, DITM is applied solely during training and fully removed at inference, incurring zero additional inference overhead. Instead of redesigning the enhancement backbone or increasing inference-time complexity, DITM improves supervised feed-forward LLIE models by reshaping the model prediction during training in a lightweight, plug-and-play, and architecture-agnostic manner. This design is particularly suitable for efficient visual enhancement systems, since it improves the training behavior of existing LLIE backbones without modifying their deployed inference architectures. As a result, the enhanced models retain the same number of inference parameters, operations, and latency as the original baselines, which is desirable for practical low-light image/video processing applications. \textcolor{black}{Of note, instead of redesigning the enhancement backbone or increasing inference-time complexity, DITM introduces a training-only interpolation process between the raw model prediction and its brightness-aligned image. The same reconstruction loss function is retained, but it is evaluated on the interpolated prediction rather than directly on the raw prediction. Therefore, DITM does not explicitly multiply samples, loss terms, or gradients by adaptive weights, nor does it introduce auxiliary prediction heads, intermediate-layer supervision objectives, or a sample-difficulty schedule. Its key distinction lies in changing the prediction involved in supervision through brightness-aware interpolation, rather than changing the loss formulation or the training schedule. Since DITM is removed after training, the enhanced models preserve the same deployed architecture, inference parameters, operations, and latency as the original baselines, making it suitable for efficient low-light image/video enhancement systems.}

Our main contributions are summarized as follows:
\begin{itemize}
\item We identify a training-stage optimization imbalance in supervised LLIE, where global brightness discrepancy tends to stabilize earlier than residual artifact suppression and fine-detail restoration.
\item We propose DITM, a \textcolor{black}{lightweight training-only module} for supervised feed-forward LLIE training that adaptively reshapes the model prediction through brightness-aware interpolation, without modifying the backbone architecture \textcolor{black}{or introducing additional learnable parameters}.
\item DITM is entirely removed at inference and improves fidelity, perceptual quality, and no-reference assessment scores across CNN-, Transformer-, and Mamba-based supervised feed-forward LLIE backbones, with zero additional inference cost.
\end{itemize}

\section{Related Work}

\subsection{Learning-based Low-Light Image Enhancement}
Existing learning-based LLIE methods mainly improve enhancement performance from two directions: network design and loss function design. The former focuses on strengthening the restoration capacity of the model itself, while the latter improves optimization through carefully designed training constraints.

From the perspective of model architecture, one major direction is to improve representation capacity. Early methods mainly relied on CNNs to address low-light degradation \cite{wei2018deep,liu2021retinex_unrolling,zhang2019kindling}, but their locality-driven design limits the modeling of long-range dependencies and global contextual relationships. To enhance long-range dependency capture, transformer-based architectures introduced self-attention for stronger non-local interaction and contextual representation \cite{cai2023retinexformer,pan2024illumination,wang2023llformer}. To reduce the computational burden of self-attention while preserving global dependency modeling, state-space models (SSMs) have recently been introduced as a more efficient alternative with linear complexity \cite{gu2023mamba,bai2024retinexmamba,zou2024wave,weng2024mamballie}. \textcolor{black}{In parallel, diffusion-based models have also been explored for LLIE, leveraging iterative denoising to improve restoration quality under severe low-light degradation \cite{zhao2023pydiff,hou2023global,jiang2023low,TCSVT2026,shang2024multi}. Related diffusion-based restoration studies have also investigated pretrained text-to-image models for generic image restoration \cite{wang2025adapting}.} Beyond architectural advancement, another important direction is explicit degradation modeling. Unlike generic restoration networks, these methods explicitly account for the complex degradation patterns in low-light images, such as noise amplification and structural corruption. By incorporating designs tailored to degradation characteristics, these approaches aim to enable more robust restoration under challenging illumination conditions \cite{xu2022snr,yan2025hvi,xu2023low,liang2024towards,feijoo2025darkir,TCSVT2026-YUV}. Another important direction explores frequency representations for LLIE. By complementing spatial features with frequency information, these methods better preserve edges, textures, and structural consistency during restoration \cite{wang2023fourllie,zhang2025prnet,zhang2024dmfourllie,TCSVT2024-Attention}. Despite their effectiveness, these architecture-driven approaches may increase computational cost or inference latency, which motivates training-side enhancement strategies that preserve the deployed inference pipeline.

Loss function design provides an efficient way to improve enhancement performance during training. For LLIE, the loss functions typically combine multiple complementary constraints to encourage structure preservation and perceptual fidelity, thereby implicitly coordinating multiple degradation factors during restoration \cite{dong2025towards,jiang2021enlightengan,guo2020zero,jiang2024lightendiffusion,wang2024zero,TCSVT2024-Gamma}. While these methods can improve optimization stability, the relative influence of different objectives is usually statically predefined and often requires extensive empirical tuning, which limits adaptability to dynamically varying degradation characteristics during training.

\textcolor{black}{Beyond low-light image enhancement, low-light video enhancement has also attracted increasing attention due to its importance in real-world dynamic scenes~\cite{li2021lowlight_survey}. Different from image-based methods that mainly focus on spatial restoration within individual frames, video enhancement further requires temporal consistency across consecutive frames to suppress motion-induced artifacts~\cite{IJCV2024-Video,TCSVT2024-video}. More recently, this research direction has been extended into event-guided low-light video enhancement~\cite{PAMI2026-Event}, where event cameras provide complementary high-dynamic-range temporal cues for robust enhancement in challenging low-light scenarios.}

\subsection{\textcolor{black}{Training-Side Strategies for Efficient Vision Models}}

\textcolor{black}{Training-side strategies have become a practical direction for efficient vision models, since they can improve representation learning, optimization behavior, or deployment efficiency without fully redesigning the backbone network. In the broader context of efficient vision model design, one related line of studies introduces lightweight structural modules into existing architectures. For example, SENet recalibrates channel-wise feature responses through a compact squeeze-and-excitation operation~\cite{CVPR2018-SENet}, while CBAM and ECA further exploit spatial-channel attention or efficient local cross-channel interaction to strengthen feature representation with limited extra complexity~\cite{woo2018cbam,wang2020eca}. Similar modular designs have been widely used in recognition, detection, segmentation, and restoration tasks to enhance multi-scale perception, spatial interaction, or task-specific feature adaptation~\cite{rao2022hornet,hashmi2023featenhancer,zhu2023denoising,TCSVT2024-OneLayer}. Although these structural modules are effective, many of them still become part of the deployed inference model, which is different from training-only strategies that aim to improve learning while preserving the original inference pipeline.}

Beyond structural plug-ins, training-time auxiliary modules have been increasingly adopted to improve learning dynamics without adding inference cost. For example, YOLOv7~\cite{wang2023yolov7} introduces auxiliary supervision branches and optimization regularizers that are active only during training and removed at inference. Similar training-time strategies have also been explored in video understanding tasks such as action recognition, where temporary training branches enhance motion representation while preserving single-stream inference~\cite{nugroho2024flow}. In the LLIE domain, some methods also employ training-time auxiliary modules to accelerate convergence and enhance feature learning, while introducing no additional computational overhead during inference. For example, FDANet~\cite{FDANet_TIP2025} introduces a raw supervision branch during training to guide feature representation, which is discarded at test time. These methods demonstrate that auxiliary pathways can provide optimization guidance while maintaining a lightweight inference pipeline, achieving a balance between improved learning and computational efficiency.

\textcolor{black}{DITM is also related to training-time optimization strategies such as auxiliary supervision, curriculum learning, self-paced learning, and adaptive loss reweighting. Deeply supervised networks introduce companion objectives at intermediate layers to improve gradient propagation and feature learning~\cite{Deeply-Supervised-Nets}, while curriculum learning adjusts the training order or difficulty of samples to guide optimization~\cite{ICML2009curriculum}. In LLIE, Luo et al.~\cite{TMM2025_SelfPaceLLIE} introduced self-paced learning into unsupervised enhancement by progressively selecting reliable samples for optimization, and Kim et al.~\cite{SPL2025} designed an intrinsic decomposition-based curriculum to guide LLIE training from easier to harder degradation patterns. Adaptive reweighting methods, such as focal loss and GradNorm, modify the relative contribution of samples, tasks, or gradients during training~\cite{ICCV2017_FocalLoss,ICML2018_GradNorm}. In image reconstruction, focal frequency loss further encourages models to focus on hard-to-reconstruct frequency components through dynamic frequency-domain weighting~\cite{ICCV2021_FocalLoss}. Different from these methods, DITM does not introduce auxiliary objectives, alter the sample curriculum, select training samples, or explicitly assign weights to loss terms. Instead, it reshapes the supervised prediction itself through brightness-aware interpolation before loss computation, thereby changing the effective training signal while keeping the original loss form and inference model unchanged.}

\textcolor{black}{Overall, existing training-side strategies improve optimization mainly by adding auxiliary branches or objectives, scheduling the learning difficulty, selecting samples progressively, or dynamically reweighting samples, tasks, gradients, or loss components. In contrast, DITM follows a different route: it reshapes the supervised prediction without introducing learnable parameters by interpolating the model prediction with a mean brightness-aligned counterpart. This design is specifically motivated by the training dynamics of supervised LLIE, where global brightness discrepancy often stabilizes earlier than residual noise and structural artifacts. Therefore, DITM is not a general loss-weighting or curriculum strategy, but a task-oriented prediction reshaping mechanism that weakens the dominance of global luminance mismatch and makes residual local degradation more influential during later optimization. Since DITM introduces no auxiliary branch, trainable parameters, auxiliary prediction, sample-selection schedule, or inference-time computation, it keeps the deployed enhancement model unchanged.}

\section{Method}

This section presents the proposed Dynamic Interpolation Training-only Module (DITM) for efficient supervised LLIE training. Section~\ref{sec:3.1} first gives an overview of the overall training framework and illustrates how DITM is integrated into the baseline network. Section~\ref{sec:3.2} introduces the construction of the mean brightness-aligned image and the dynamically interpolated prediction. Section~\ref{sec:3.3} introduces the design of the adaptive interpolation factor $s$ under a Weighted Least Squares (WLS) perspective. Section~\ref{sec:ditm} offers a qualitative analysis of the training response induced by DITM. \textcolor{black}{Section~\ref{sec:consistency} further discusses training–inference consistency}. Section~\ref{sec:3.5} discusses the computational overhead introduced by DITM.

\subsection{Overview}
\label{sec:3.1}
\begin{figure*}[t]
  \centering
  \includegraphics[width=\linewidth]{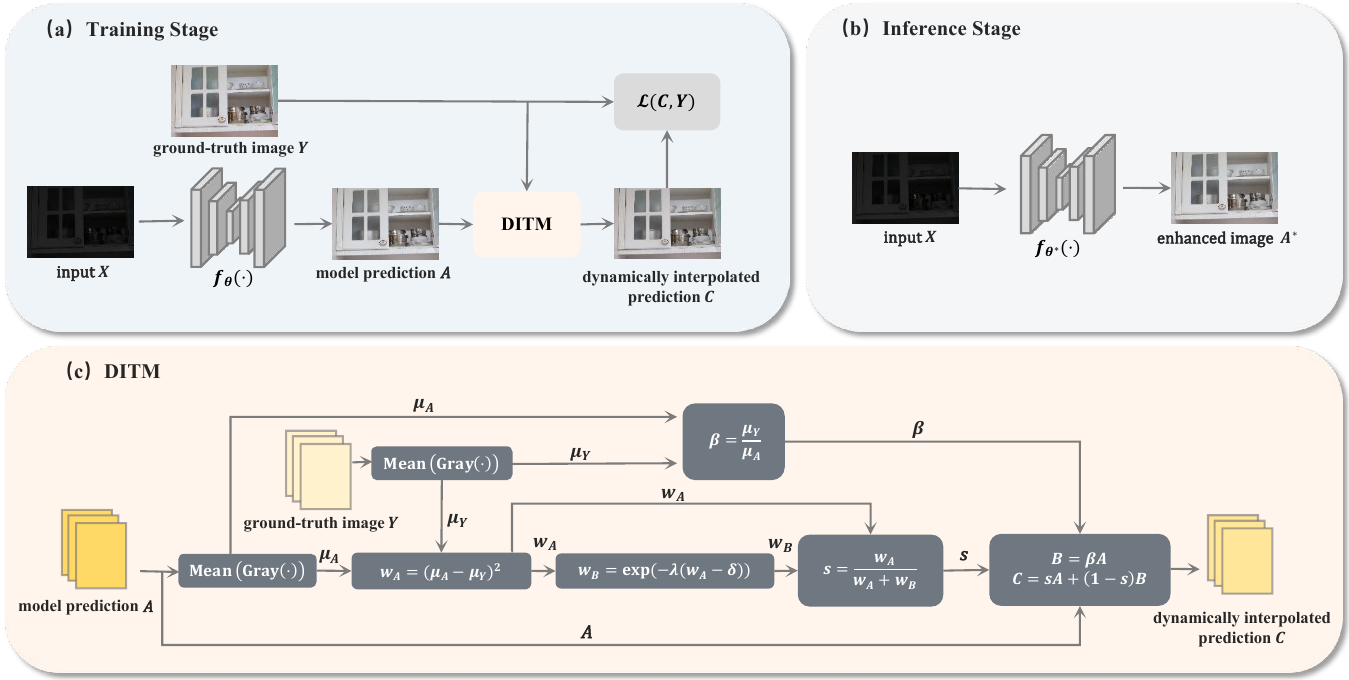}
  \caption{Training and inference pipeline of the proposed DITM. (a) Training Stage: DITM adjusts the model prediction $A$ of the baseline model $f_\theta(\cdot)$ by leveraging the global mean brightness values of the ground-truth image $Y$, generating an interpolated prediction $C$ for loss computation. (b) Inference Stage: DITM is discarded, and the trained baseline $f_{\theta^*}(\cdot)$ produces the final enhanced image $A^*$ with no additional inference-time computation. (c) DITM: This module adaptively computes an interpolation factor $s$ based on the brightness discrepancy between $A$ and $Y$ to yield the adjusted prediction $C$.}
  \label{fig:jiegoutu}
\end{figure*}

To improve supervised LLIE models without increasing inference complexity, we propose a plug-and-play training-only module, termed the Dynamic Interpolation Training-only Module (DITM). As illustrated in Fig.~\ref{fig:jiegoutu}(a), the overall training framework consists of two components: the baseline enhancement network $f_\theta(\cdot)$ and DITM. Instead of modifying the internal architecture of the backbone, DITM operates on the model prediction $A$ and regulates the optimization process during training. Owing to its architecture-agnostic design, DITM can be integrated into diverse supervised LLIE backbones while leaving the inference architecture unchanged. 

Given a low-light input image $X$, the enhancement network $f_\theta(\cdot)$ produces a prediction $A = f_\theta(X)$. Instead of directly using $A$ for loss computation, DITM leverages the global mean brightness of the ground-truth image $Y$ and transforms $A$ into a dynamically interpolated prediction $C$ via:

\begin{equation}
C = \text{DITM}(A, Y).
\end{equation}

In essence, $C$ adaptively reshapes the model prediction $A$. The internal operation of DITM is illustrated in Fig.~\ref{fig:jiegoutu}(c). $C$ serves as a dynamically interpolated prediction that encourages the optimization process to better balance global luminance restoration, local artifact suppression, and detail restoration during training. The network parameters $\theta$ are then optimized by minimizing the loss between the interpolated prediction $C$ and the ground-truth image $Y$:
\begin{equation}
\theta^* = \arg\min_\theta \mathcal{L}(C, Y).
\end{equation}

It is worth emphasizing that DITM is used only during training and is removed during inference. As illustrated in Fig.~\ref{fig:jiegoutu}(b), the baseline enhancement network directly produces the final enhanced image $A^* = f_{\theta^*}(X)$. \textcolor{black}{Although supervision is applied to the interpolated prediction $C$, DITM introduces no learnable parameters, and all gradients propagate through $A$ to update only the baseline enhancement network $f_\theta(X)$.} Consequently, the trained baseline enhancement network $f_{\theta^*}(X)$ achieves improved restoration performance during inference without introducing additional parameters, operations, or latency at inference. 

\subsection{Dynamic Construction of Interpolated Prediction}
\label{sec:3.2}
\textbf{Mean Brightness-Aligned Prediction.}
Using the mean-brightness ratio between $Y$ and $A$, we construct $B$ by linearly rescaling $A$:
\begin{equation}
B = \beta  A, \quad \text{where } \beta = \frac{\mu_Y}{\mu_A} = \frac{\text{Mean}(\text{Gray}(Y))}{\text{Mean}(\text{Gray}(A))}.
\end{equation}
Here, $\mu_Y$ and $\mu_A$ denote the global mean brightness values of the ground-truth image $Y$ and the model prediction $A$, respectively. {\color{black}They are obtained by spatially averaging the grayscale images converted from RGB using the weights $(0.2989, 0.5870, 0.1140)$. Since $B$ is constructed by uniformly scaling all pixels of $A$ with the same positive factor $\beta$, this operation changes only its global intensity scale without altering pixel locations or spatial relationships. Consequently, $B$ matches the global mean brightness of $Y$ while retaining the structural layout, edges, and local content inherited from $A$. It should be noted that this operation provides global mean-brightness alignment rather than explicit spatial illumination modeling.}

\textbf{Dynamic Prediction Interpolation.} With the mean brightness-aligned image $B$ established, the dynamically interpolated prediction $C$ is constructed through adaptive interpolation between $A$ and $B$:
\begin{equation}
\label{eq:4}
C = s  A + (1-s)  B, \quad s \in (0,1),
\end{equation}
{\color{black}Here, $s$ is a sample-wise adaptive interpolation factor determined by the current brightness discrepancy, which dynamically controls the balance between $A$ and $B$. In our implementation, $s$ is detached before being used to construct $C$. Therefore, its value varies across samples and training iterations but is treated as a constant during backpropagation to avoid introducing an additional optimization path through the factor itself.}

Instead of directly supervising the model prediction $A$, the proposed module applies supervision to the interpolated prediction $C$. During training, network parameters are updated by minimizing the discrepancy between $C$ and the ground-truth image $Y$. This interpolation introduces brightness-aware prediction reshaping into the supervised training objective. Instead of directly imposing local structural constraints, DITM weakens the dominance of global brightness mismatch in the reconstruction loss. Consequently, when global luminance has been largely corrected, residual local errors, such as noise-related deviations and structural artifacts, become relatively more influential during backpropagation. The effectiveness of DITM therefore depends on how this interpolation strength is adaptively controlled during training. Because different training stages and image states may require different interpolation strengths, the factor $s$ is designed to provide per-sample adjustment and maintain a suitable balance throughout optimization.

\subsection{Adaptive Interpolation Factor Design}
\label{sec:3.3}

As a key component of DITM, the adaptive interpolation factor $s$ controls the balance between $A$ and $B$ during training. To achieve effective adjustment of $s$, its design follows two core principles:

\begin{itemize}
\item Brightness-Discrepancy Awareness: $s$ should reflect the brightness discrepancy between the model prediction $A$ and the ground-truth image $Y$, so as to provide an adaptive signal that guides the balance between $A$ and $B$ during training.

\item Smooth Optimization-Focus Shift: $s$ should decay smoothly as the brightness discrepancy decreases, enabling the optimization to gradually shift focus toward refining image degradations beyond global luminance.

\end{itemize}

\textbf{Weighted Optimization Framework.} Adaptive interpolation essentially aims to strike a reasonable balance between $A$ and $B$. To provide an analytical interpretation of this balance, we describe the interpolation process from a Weighted Least Squares perspective. In this formulation, the interpolated prediction $C$ is treated as the weighted centroid between $A$ and $B$, obtained by minimizing the following weighted interpolation objective:

\begin{equation}
\mathcal{J}_C(C; w_A, w_B) = \frac{1}{2} w_A \|C - A\|^2 + \frac{1}{2} w_B \|C - B\|^2,
\end{equation}
where $w_A$ and $w_B$ are non-negative weights that control the relative contributions of $A$ and $B$ in the construction of $C$ during training. This formulation is used to interpret the interpolated prediction $C$ as a weighted centroid between $A$ and $B$. Under this framework, the resulting solution naturally balances the influence of $A$ and $B$, such that $C$ approaches $A$ when $w_A$ dominates and approaches $B$ when $w_B$ dominates. Under this view, minimizing the weighted interpolation objective yields a weighted combination of $A$ and $B$, which is consistent with the interpolation form in Eq.~(\ref{eq:4}).

By setting the derivative $\frac{\partial \mathcal{J}_C}{\partial C} = 0$, we obtain the closed-form solution to this weighted interpolation problem:
\begin{equation}
\label{eq:6}
C_{\text{opt}} = \frac{w_A A + w_B B}{w_A + w_B}.
\end{equation}
Equivalently, according to Eq.~(\ref{eq:4}), we write $C_{\text{opt}} = s A + (1-s) B$ with $s = \frac{w_A}{w_A + w_B}$.

\textbf{Brightness Discrepancy Design of Weight $w_A$.}
To satisfy the Brightness-Discrepancy Awareness, the weight $w_A$ is designed to quantify the brightness discrepancy between $A$ and $Y$. To this end, the global brightness discrepancy between $A$ and $Y$ is measured based on the mean intensities $\mu_A$ and $\mu_Y$. Accordingly, the weight $w_A$ is defined as
\begin{equation}
w_A = (\mu_A - \mu_Y)^2.
\end{equation}

The quadratic form emphasizes larger brightness deviations while providing a smooth adaptive weighting signal. In our implementation, both $A$ and $Y$ are normalized to [0, 1] before computing $\mu_A$, $\mu_Y$, and $w_A$. Therefore, $w_A$ is defined on a bounded normalized brightness-discrepancy scale. Since the adaptive factor $s$ is detached before backpropagation, $w_A$ is used to determine the interpolation strength in the forward pass rather than to introduce an additional optimizable gradient pathway.

\textbf{Complementary Design of Weight $w_B$.} 
To implement the Smooth Optimization-Focus Shift, the weight $w_B$ is designed as a non-linear complementary function of $w_A$. To achieve this, we introduce an exponential form with a threshold mechanism to control the transition speed:
\begin{equation}
\label{eq:8}
w_B = \exp(-\lambda (w_A - \delta)), \quad \lambda > 0, \ \delta > 0,
\end{equation}
where $\lambda$ and $\delta$ are \textcolor{black}{preset hyperparameters} controlling the transition sharpness and the brightness-discrepancy transition point of the complementary weighting, respectively. Since $w_A$ is defined on the normalized squared mean-brightness discrepancy scale, $\delta$ is specified on the same scale and is thus comparable across different datasets and backbone networks. In this paper, $\lambda$ and $\delta$ are empirically fixed across all datasets and backbone networks, and their influence is further analyzed in the hyperparameter study. This exponential formulation provides a smooth monotonic response to the brightness-discrepancy weight, enabling the interpolation factor to change more sensitively when $w_A$ approaches the transition region. The interpolation factor $s$ is then defined as

\begin{equation}
\label{eq:9}
s = \frac{w_A}{w_A + \exp(-\lambda (w_A - \delta))},
\end{equation}
\textcolor{black}{the brightness statistics and $s$ are computed independently for each sample, without batch-wise aggregation.}

Overall, this design defines a brightness-aware adaptive interpolation mechanism. Empirically, it allows the interpolation factor $s$ to vary with the brightness discrepancy and to maintain a suitable balance throughout training. The weight $w_A$ quantifies the brightness discrepancy between $A$ and $Y$ and diminishes as $A$ gets closer in brightness to $Y$. The exponential complementary function defined in Eq.~(\ref{eq:8}) transforms the decreasing discrepancy in $w_A$ into a smoothly increasing $w_B$, with $\lambda$ controlling the transition rate and $\delta$ specifying the onset of accelerated change. Importantly, this function induces a gradual adjustment of the interpolation factor $s$. When $w_A$ is above the transition point $\delta$, $s$ changes relatively mildly and retains a larger contribution from the original prediction $A$. When $w_A$ falls below this transition point, $s$ decreases more sensitively, thereby increasing the relative contribution of the brightness-aligned image $B$. As a result, this design enables $s$ to be adjusted according to different levels of brightness discrepancy, maintaining an appropriate interpolation balance throughout training. 

In practice, during the early training stage, the model prediction $A$ usually presents a relatively large brightness discrepancy from the ground-truth image $Y$. Accordingly, $w_A$ tends to be larger on the normalized brightness-discrepancy scale, which preserves a relatively large interpolation factor $s$ and retains a stronger contribution from the original prediction $A$ at this stage. As training progresses and the brightness discrepancy gradually decreases, $w_A$ correspondingly declines, causing $s$ to decrease and increasing the relative contribution of the brightness-aligned image $B$. In this stage, the global brightness discrepancy contributes less dominantly to the reconstruction loss, so residual local discrepancies, including noise-related deviations and structural artifacts, exert relatively stronger influence on the gradient update. 

\subsection{Analysis of Training Response Induced by DITM}
\label{sec:ditm}

\textcolor{black}{To examine the training behavior induced by DITM, we analyze the response of the training objective in the input space. As defined in Eq.~(\ref{eq:4}), $C$ is obtained by adaptively interpolating $A$ and $B$ with the factor $s$. Although $s$ is dynamically recomputed for each sample at every training iteration, its gradient is stopped before backpropagation. Therefore, within each optimization step, $s$ is treated as a constant and does not introduce an additional gradient path through its own computation. Accordingly, in the following derivation, $s$ denotes the value computed during the forward pass, with its gradient stopped during backpropagation.}

Let $\mathcal{L}$ denote the reconstruction loss imposed on the interpolated prediction $C$. Although the input image $X$ is not an optimization variable, the derivative of the loss with respect to $X$ can serve as a local response indicator of how the objective behaves across different spatial regions. The corresponding response can be written as

\begin{equation}
\frac{\partial \mathcal{L}}{\partial X}
=
\frac{\partial \mathcal{L}}{\partial C}
\left(
s
\frac{\partial A}{\partial X}
+
\left(
1-s
\right)
\frac{\partial B}{\partial X}
\right).
\label{eq:response1}
\end{equation}
Since $B=\beta A$, its derivative with respect to $X$ can be further decomposed as
\begin{equation}
\frac{\partial B}{\partial X}
=
\beta
\frac{\partial A}{\partial X}
+
A
\frac{\partial \beta}{\partial X}.
\label{eq:response2}
\end{equation}
Substituting this into the previous equation gives
\begin{equation}
\frac{\partial \mathcal{L}}{\partial X}
=
\frac{\partial \mathcal{L}}{\partial C}
\left[
\left(s+(1-s)\beta\right)\frac{\partial A}{\partial X}
+
(1-s)A\frac{\partial \beta}{\partial X}
\right].
\label{eq:response3}
\end{equation}

The above formulation reveals that the training response consists of two components. The first term is proportional to $\frac{\partial A}{\partial X}$ and preserves the spatially varying gradients produced by the backbone network. The second term arises from $\frac{\partial \beta}{\partial X}$, where $\beta$ depends on the global mean brightness $\mu_A$. Since $\mu_A$ aggregates information over the entire prediction, this term introduces a luminance-related global coupling effect into the response. Nevertheless, this term does not impose an explicit local structural constraint; rather, its spatial effect is indirectly shaped by the prediction $A$ and the backbone Jacobian.

At the early training stage, $s$ remains relatively large, so the training objective places greater weight on the prediction $A$. In this regime, the response is mainly associated with correcting large global luminance errors and therefore tends to be broadly distributed rather than clearly localized to specific residual structures. As training proceeds and the brightness discrepancy decreases, $s$ gradually decays according to the adaptive interpolation rule in Section~\ref{sec:3.3}, increasing the relative influence of the brightness-aligned image $B$. Because $B$ reduces the relative dominance of global brightness mismatch in the reconstruction loss, residual local discrepancies become more influential in the resulting gradient response.

Consequently, under DITM, the response tends to become less dominated by global luminance correction and more aligned with spatially non-uniform residual errors, including structural boundaries, texture regions, and noise-contaminated areas. This pattern is consistent with the view that DITM makes noise-related and structure-related residuals relatively more visible to the training objective during the middle and late training stages. The corresponding experimental observations in Section~\ref{sec:4:4} further provide empirical support for this training behavior.

{\color{black}
\subsection{Further Discussion on Training--Inference Consistency}
\label{sec:consistency}

To further clarify the training–inference consistency of DITM, we discuss the role of the ground-truth brightness statistic during training and its relation to the deployed inference model. During training, the backbone receives only the low-light image $X$ and produces the raw model prediction $A=f_{\theta}(X)$. The ground-truth image $Y$ is used only on the loss side to construct the interpolated prediction $C=\mathrm{DITM}(A,Y)$, and is not provided as an input, auxiliary condition, or additional feature to the enhancement network. The corresponding training objective is $\mathcal{L}(C,Y)$.

DITM introduces neither learnable parameters nor an additional prediction branch. Although the loss is imposed on $C$, the backbone parameters are updated through the dependency of $C$ on the raw prediction $A$:

\begin{equation}
\frac{\partial \mathcal{L}(C,Y)}{\partial \theta}
=
\frac{\partial \mathcal{L}}{\partial C}
\left[
sI
+
(1-s)
\frac{\partial B}{\partial A}
\right]
\frac{\partial A}{\partial \theta}.
\end{equation}
Here, $I$ denotes the identity operator. In our implementation, the adaptive interpolation factor $s$ is detached before backpropagation and is therefore treated as a constant for each sample within the current optimization step. Consequently, the ground-truth brightness statistic affects the optimization signal only through the deterministic loss-side transformation that reshapes $A$ into $C$, rather than through a learnable GT-dependent control pathway.

Importantly, DITM uses only the scalar mean-brightness statistic $\mu_Y$ from the ground-truth image. Since $B={\beta}A$, the interpolated prediction can be equivalently written as $C=[s+(1-s)\beta]A$. Therefore, $C$ does not borrow any spatial content, such as structures or textures, from $Y$; all its spatial information is inherited from the raw prediction $A$. Consequently, improvements in local detail restoration and artifact suppression are still learned by the backbone and reflected in $A$.

After training, DITM is completely removed, and the deployed model directly produces $A^{*}=f_{\theta^{*}}(X)$. Therefore, the inference dependency remains $X\rightarrow A^{*}$, and neither the ground-truth image nor its brightness statistics are required at test time. DITM changes how the supervised training signal is presented to the backbone while preserving the original feed-forward inference mapping. All reported validation and test results are computed from the raw output $A^{*}$, rather than from the GT-dependent interpolated prediction $C$; when available, the ground-truth image is used only for metric computation.
}

\subsection{Computational Complexity of DITM}
\label{sec:3.5}
Based on Fig.~\ref{fig:jiegoutu}, we analyze the additional computational and memory overhead introduced by DITM. The main computations involve constructing the mean brightness-aligned image $B$ and the interpolated prediction $C$. Computing $B$ requires evaluating the mean brightness of the model prediction $A$, which has linear complexity $\mathcal{O}(H\times W)$ for an image of size $H \times W$. Generating $C$ involves element-wise interpolation between $A$ and $B$, which is also linear in $H \times W$. The adaptive interpolation weight $s$ is computed from scalar statistics and incurs negligible cost. Consequently, the overall additional computational complexity introduced by DITM remains $\mathcal{O}(H \times W)$.

In terms of memory, DITM requires storing $B$ and $C$, each of size $H \times W$. Compared to the backbone, whose dominant cost arises from multi-channel feature maps, both the extra computational and memory overheads introduced by DITM are negligible. Moreover, no additional trainable parameters are introduced, and the inference-time architecture and complexity remain identical to those of the original backbone.

\section{Experiments}
\label{sec:4}
\textcolor{black}{
This section experimentally evaluates DITM. Section~\ref{sec:experimental_setup} introduces the datasets, evaluation metrics, baseline models, and implementation details. Sections~\ref{sec:quantitative_evaluation} and~\ref{sec:visual_evaluation} present quantitative and qualitative comparisons, respectively. Section~\ref{sec:4:4} further analyzes alternative interpolation strategies, hyperparameters, training behavior and overhead, numerical robustness under extremely low-light conditions, and performance under spatially non-uniform illumination. Finally, Section~\ref{sec:applicability} discusses the applicability scope and limitations of DITM.
}

\subsection{Experimental Setup}
\label{sec:experimental_setup}
\textbf{Datasets.} We evaluated our proposed DITM on eight widely used low-light image enhancement datasets. The paired datasets include LOLv1 \cite{wei2018deep}, LOLv2-real \cite{yang2021sparse}, and LOLv2-synthetic \cite{yang2021sparse}, which provide aligned low-light and normal-light images for supervised evaluation. LOLv1 contains 485 training images and 15 testing images, LOLv2-real includes 689 training images and 100 testing images, and LOLv2-synthetic provides 900 training images and 100 testing images. For unpaired datasets, we used LIME \cite{guo2017lime}, VV \cite{vonikakis2018evaluation}, NPE \cite{wang2013naturalness}, MEF \cite{ma2015perceptual}, and DICM \cite{lee2013contrast}. 

\textbf{Baseline Methods.} To evaluate the generality of DITM across different enhancement architectures, we integrated it into seven models covering three major architectural paradigms: CNN-based, Transformer-based, and Mamba-based approaches. The CNN-based baselines include RetinexNet \cite{wei2018deep}, MIRNet \cite{zamir2020mirnet}, and CID-Net \cite{yan2025hvi}; the Transformer-based baselines include LLFormer \cite{wang2023llformer} and RetinexFormer \cite{cai2023retinexformer}; and the Mamba-based baselines include RetinexMamba \cite{bai2024retinexmamba} and MambaLLIE \cite{weng2024mamballie}. 

\textbf{Evaluation Metrics.} To evaluate both distortion fidelity and perceptual quality, we adopted a combination of full-reference and no-reference metrics. For paired datasets, we employed standard full-reference metrics including PSNR, SSIM, DeepSSIM \cite{deepssimPAMI}, and LPIPS \cite{zhang2018unreasonable} to evaluate pixel-wise fidelity and perceptual similarity. For unpaired datasets without ground-truth references, we followed the evaluation protocol in~\cite{zhang2025augmenting} and adopted perceptually-aligned no-reference metrics that better reflect human visual perception. Specifically, we used MUSIQ~\cite{ke2021musiq} and Q-Align~\cite{wu2024qalign}, the latter of which provides two complementary evaluation scores: image quality assessment (IQA) and image aesthetic assessment (IAA). These perceptually aligned no-reference metrics, including the LMM-based Q-Align, provide complementary evaluation of perceptual quality and aesthetic appeal for unpaired low-light images. It is worth noting that we also reported Q-Align (IQA and IAA) on paired datasets for complementary perceptual evaluation.

{\color{black}
\textbf{Implementation Details.} We integrated DITM into each baseline and retrained both the original and DITM-augmented models from scratch following the official configurations. For each paired comparison, all training and evaluation settings were kept identical, including the dataset, preprocessing, patch size, batch size, optimizer, learning rate and its schedule, training length, data augmentation, loss function, and evaluation protocol. The only difference was whether DITM was enabled during training, while the inference procedure remained unchanged. Each baseline and its DITM-augmented counterpart were independently trained three times, and the results are reported as mean and standard deviation. The supervised feed-forward LLIE experiments were conducted using NVIDIA RTX 4090 GPUs. The hyperparameters $\lambda$ and $\delta$ were fixed to 10 and 0.3, respectively, across all backbones. All brightness statistics were computed on images normalized to $[0,1]$, and $\delta$ is therefore defined on the normalized squared mean-brightness discrepancy scale.
}

\begin{table*}[t]
\caption{\textcolor{black}{Quantitative comparison on LOL datasets. Each setting was independently repeated three times, and the mean and standard deviation are reported. Values in parentheses denote the mean change of DITM over the corresponding baseline. Red indicates improvement, while green indicates degradation.}}
\label{tab:paired_results}
\centering
\scriptsize
\setlength{\tabcolsep}{2.6pt}
\resizebox{\textwidth}{!}{
\begin{tabular}{>{\centering\arraybackslash}m{2.1cm}|>{\raggedright\arraybackslash}p{3.0cm}|cccccc}
\toprule
\multicolumn{1}{c|}{\multirow{2}{*}{\textbf{Datasets}}} 
& \multicolumn{1}{c|}{\multirow{2}{*}{\textbf{Methods}}} 
& \multicolumn{6}{c}{\textbf{Evaluation Metrics}} \\
& & PSNR$\uparrow$ & SSIM$\uparrow$ & LPIPS$\downarrow$ & DeepSSIM$\uparrow$ & IQA$\uparrow$ & IAA$\uparrow$ \\
\midrule

\multirow{14}{*}{LOLv1} 
& RetinexNet 
& 17.076$\pm$0.080 & 0.7095$\pm$0.0020 & 0.424$\pm$0.004 & 0.3834$\pm$0.0028 & 2.032$\pm$0.019 & 1.204$\pm$0.007 \\
& RetinexNet with DITM 
& 17.546$\pm$0.111 \gainpos{+0.470} & 0.7190$\pm$0.0013 \gainpos{+0.0095} & 0.403$\pm$0.015 \gainpos{-0.021} & 0.4002$\pm$0.0208 \gainpos{+0.0168} & 2.144$\pm$0.036 \gainpos{+0.112} & 1.279$\pm$0.047 \gainpos{+0.075} \\
\cmidrule(lr{1em}){2-8}

& MIRNet 
& 21.447$\pm$0.107 & 0.7849$\pm$0.0030 & 0.225$\pm$0.001 & 0.6072$\pm$0.0032 & 2.937$\pm$0.016 & 1.759$\pm$0.020 \\
& MIRNet with DITM 
& 21.703$\pm$0.101 \gainpos{+0.256} & 0.7963$\pm$0.0011 \gainpos{+0.0114} & 0.217$\pm$0.004 \gainpos{-0.008} & 0.6265$\pm$0.0043 \gainpos{+0.0193} & 2.973$\pm$0.030 \gainpos{+0.036} & 1.750$\pm$0.019 \gainneg{-0.009} \\
\cmidrule(lr{1em}){2-8}

& CID-Net 
& 24.102$\pm$0.184 & 0.8563$\pm$0.0016 & 0.098$\pm$0.003 & 0.8193$\pm$0.0125 & 3.889$\pm$0.048 & 2.097$\pm$0.009 \\
& CID-Net with DITM 
& 24.447$\pm$0.372 \gainpos{+0.345} & 0.8628$\pm$0.0009 \gainpos{+0.0065} & 0.087$\pm$0.001 \gainpos{-0.011} & 0.8494$\pm$0.0042 \gainpos{+0.0301} & 3.969$\pm$0.013 \gainpos{+0.080} & 2.125$\pm$0.003 \gainpos{+0.028} \\
\cmidrule(lr{1em}){2-8}

& LLFormer 
& 23.236$\pm$0.161 & 0.8129$\pm$0.0011 & 0.171$\pm$0.002 & 0.6632$\pm$0.0019 & 3.125$\pm$0.025 & 1.839$\pm$0.019 \\
& LLFormer with DITM 
& 23.356$\pm$0.191 \gainpos{+0.120} & 0.8215$\pm$0.0018 \gainpos{+0.0086} & 0.162$\pm$0.003 \gainpos{-0.009} & 0.6763$\pm$0.0038 \gainpos{+0.0131} & 3.176$\pm$0.019 \gainpos{+0.051} & 1.867$\pm$0.005 \gainpos{+0.028} \\
\cmidrule(lr{1em}){2-8}

& RetinexFormer 
& 23.478$\pm$0.331 & 0.8266$\pm$0.0015 & 0.152$\pm$0.003 & 0.6739$\pm$0.0072 & 3.265$\pm$0.016 & 1.932$\pm$0.007 \\
& RetinexFormer with DITM 
& 23.946$\pm$0.152 \gainpos{+0.468} & 0.8437$\pm$0.0014 \gainpos{+0.0171} & 0.121$\pm$0.002 \gainpos{-0.031} & 0.7470$\pm$0.0092 \gainpos{+0.0731} & 3.611$\pm$0.056 \gainpos{+0.346} & 2.036$\pm$0.006 \gainpos{+0.104} \\
\cmidrule(lr{1em}){2-8}

& RetinexMamba 
& 23.345$\pm$0.117 & 0.8311$\pm$0.0027 & 0.142$\pm$0.007 & 0.7001$\pm$0.0124 & 3.364$\pm$0.047 & 1.986$\pm$0.014 \\
& RetinexMamba with DITM 
& 23.729$\pm$0.176 \gainpos{+0.384} & 0.8458$\pm$0.0008 \gainpos{+0.0147} & 0.123$\pm$0.004 \gainpos{-0.019} & 0.7536$\pm$0.0109 \gainpos{+0.0535} & 3.644$\pm$0.067 \gainpos{+0.280} & 2.038$\pm$0.011 \gainpos{+0.052} \\
\cmidrule(lr{1em}){2-8}

& MambaLLIE 
& 22.985$\pm$0.193 & 0.8232$\pm$0.0019 & 0.151$\pm$0.004 & 0.6916$\pm$0.0027 & 3.369$\pm$0.026 & 2.037$\pm$0.023 \\
& MambaLLIE with DITM 
& 23.291$\pm$0.259 \gainpos{+0.306} & 0.8436$\pm$0.0028 \gainpos{+0.0204} & 0.128$\pm$0.002 \gainpos{-0.023} & 0.7418$\pm$0.0021 \gainpos{+0.0502} & 3.574$\pm$0.047 \gainpos{+0.205} & 2.055$\pm$0.021 \gainpos{+0.018} \\

\midrule

\multirow{14}{*}{LOLv2-real} 
& RetinexNet 
& 16.494$\pm$0.437 & 0.6705$\pm$0.0123 & 0.519$\pm$0.021 & 0.3049$\pm$0.0272 & 1.826$\pm$0.013 & 1.223$\pm$0.008 \\
& RetinexNet with DITM 
& 17.206$\pm$0.354 \gainpos{+0.712} & 0.7015$\pm$0.0081 \gainpos{+0.0310} & 0.501$\pm$0.032 \gainpos{-0.018} & 0.3437$\pm$0.0140 \gainpos{+0.0388} & 2.016$\pm$0.171 \gainpos{+0.190} & 1.338$\pm$0.104 \gainpos{+0.115} \\
\cmidrule(lr{1em}){2-8}

& MIRNet 
& 21.504$\pm$0.173 & 0.8066$\pm$0.0061 & 0.324$\pm$0.035 & 0.4913$\pm$0.0411 & 2.448$\pm$0.161 & 1.439$\pm$0.092 \\
& MIRNet with DITM 
& 21.863$\pm$0.147 \gainpos{+0.359} & 0.8210$\pm$0.0033 \gainpos{+0.0144} & 0.263$\pm$0.017 \gainpos{-0.061} & 0.5436$\pm$0.0039 \gainpos{+0.0523} & 2.698$\pm$0.052 \gainpos{+0.250} & 1.597$\pm$0.015 \gainpos{+0.158} \\
\cmidrule(lr{1em}){2-8}

& CID-Net 
& 22.711$\pm$0.192 & 0.8526$\pm$0.0029 & 0.138$\pm$0.002 & 0.7821$\pm$0.0051 & 3.551$\pm$0.012 & 1.996$\pm$0.015 \\
& CID-Net with DITM 
& 23.010$\pm$0.187 \gainpos{+0.299} & 0.8613$\pm$0.0010 \gainpos{+0.0087} & 0.127$\pm$0.004 \gainpos{-0.011} & 0.7981$\pm$0.0022 \gainpos{+0.0160} & 3.616$\pm$0.018 \gainpos{+0.065} & 2.001$\pm$0.013 \gainpos{+0.005} \\
\cmidrule(lr{1em}){2-8}

& LLFormer 
& 21.178$\pm$0.035 & 0.8092$\pm$0.0053 & 0.221$\pm$0.021 & 0.5889$\pm$0.0299 & 2.802$\pm$0.099 & 1.666$\pm$0.069 \\
& LLFormer with DITM 
& 21.482$\pm$0.161 \gainpos{+0.304} & 0.8230$\pm$0.0064 \gainpos{+0.0138} & 0.193$\pm$0.009 \gainpos{-0.028} & 0.6289$\pm$0.0175 \gainpos{+0.0400} & 2.920$\pm$0.081 \gainpos{+0.118} & 1.751$\pm$0.042 \gainpos{+0.085} \\
\cmidrule(lr{1em}){2-8}

& RetinexFormer 
& 21.977$\pm$0.083 & 0.8395$\pm$0.0022 & 0.171$\pm$0.006 & 0.6645$\pm$0.0030 & 3.183$\pm$0.016 & 1.874$\pm$0.007 \\
& RetinexFormer with DITM 
& 22.641$\pm$0.254 \gainpos{+0.664} & 0.8544$\pm$0.0033 \gainpos{+0.0149} & 0.145$\pm$0.006 \gainpos{-0.026} & 0.7196$\pm$0.0034 \gainpos{+0.0551} & 3.430$\pm$0.017 \gainpos{+0.247} & 1.970$\pm$0.009 \gainpos{+0.096} \\
\cmidrule(lr{1em}){2-8}

& RetinexMamba 
& 21.960$\pm$0.367 & 0.8425$\pm$0.0069 & 0.165$\pm$0.005 & 0.6929$\pm$0.0079 & 3.295$\pm$0.073 & 1.958$\pm$0.031 \\
& RetinexMamba with DITM 
& 22.394$\pm$0.234 \gainpos{+0.434} & 0.8600$\pm$0.0022 \gainpos{+0.0175} & 0.146$\pm$0.005 \gainpos{-0.019} & 0.7461$\pm$0.0193 \gainpos{+0.0532} & 3.576$\pm$0.108 \gainpos{+0.281} & 2.011$\pm$0.023 \gainpos{+0.053} \\
\cmidrule(lr{1em}){2-8}

& MambaLLIE 
& 21.586$\pm$0.129 & 0.8486$\pm$0.0020 & 0.162$\pm$0.004 & 0.7179$\pm$0.0035 & 3.399$\pm$0.042 & 2.005$\pm$0.014 \\
& MambaLLIE with DITM 
& 22.346$\pm$0.261 \gainpos{+0.760} & 0.8562$\pm$0.0050 \gainpos{+0.0076} & 0.152$\pm$0.007 \gainpos{-0.010} & 0.7298$\pm$0.0036 \gainpos{+0.0119} & 3.408$\pm$0.017 \gainpos{+0.009} & 1.993$\pm$0.008 \gainneg{-0.012} \\

\midrule

\multirow{14}{*}{LOLv2-synthetic} 
& RetinexNet 
& 18.965$\pm$0.098 & 0.7947$\pm$0.0005 & 0.252$\pm$0.002 & 0.6126$\pm$0.0016 & 2.449$\pm$0.013 & 1.756$\pm$0.014 \\
& RetinexNet with DITM 
& 19.276$\pm$0.068 \gainpos{+0.311} & 0.8086$\pm$0.0014 \gainpos{+0.0139} & 0.238$\pm$0.005 \gainpos{-0.014} & 0.6382$\pm$0.0094 \gainpos{+0.0256} & 2.590$\pm$0.014 \gainpos{+0.141} & 1.814$\pm$0.008 \gainpos{+0.058} \\
\cmidrule(lr{1em}){2-8}

& MIRNet 
& 21.972$\pm$0.077 & 0.8935$\pm$0.0006 & 0.117$\pm$0.002 & 0.8196$\pm$0.0020 & 2.970$\pm$0.024 & 2.200$\pm$0.007 \\
& MIRNet with DITM 
& 22.098$\pm$0.092 \gainpos{+0.126} & 0.8990$\pm$0.0008 \gainpos{+0.0055} & 0.105$\pm$0.001 \gainpos{-0.012} & 0.8408$\pm$0.0012 \gainpos{+0.0212} & 3.051$\pm$0.007 \gainpos{+0.081} & 2.201$\pm$0.010 \gainpos{+0.001} \\
\cmidrule(lr{1em}){2-8}

& CID-Net 
& 25.519$\pm$0.088 & 0.9339$\pm$0.0011 & 0.049$\pm$0.000 & 0.9243$\pm$0.0020 & 3.283$\pm$0.004 & 2.163$\pm$0.004 \\
& CID-Net with DITM 
& 25.719$\pm$0.250 \gainpos{+0.200} & 0.9384$\pm$0.0015 \gainpos{+0.0045} & 0.045$\pm$0.001 \gainpos{-0.004} & 0.9296$\pm$0.0008 \gainpos{+0.0053} & 3.304$\pm$0.010 \gainpos{+0.021} & 2.176$\pm$0.005 \gainpos{+0.013} \\
\cmidrule(lr{1em}){2-8}

& LLFormer 
& 24.860$\pm$0.083 & 0.9268$\pm$0.0006 & 0.059$\pm$0.001 & 0.9030$\pm$0.0011 & 3.164$\pm$0.003 & 2.117$\pm$0.005 \\
& LLFormer with DITM 
& 25.204$\pm$0.136 \gainpos{+0.344} & 0.9327$\pm$0.0005 \gainpos{+0.0059} & 0.054$\pm$0.001 \gainpos{-0.005} & 0.9134$\pm$0.0007 \gainpos{+0.0104} & 3.192$\pm$0.003 \gainpos{+0.028} & 2.128$\pm$0.003 \gainpos{+0.011} \\
\cmidrule(lr{1em}){2-8}

& RetinexFormer 
& 25.308$\pm$0.173 & 0.9282$\pm$0.0013 & 0.060$\pm$0.001 & 0.9061$\pm$0.0010 & 3.182$\pm$0.007 & 2.130$\pm$0.002 \\
& RetinexFormer with DITM 
& 25.657$\pm$0.101 \gainpos{+0.349} & 0.9367$\pm$0.0025 \gainpos{+0.0085} & 0.052$\pm$0.002 \gainpos{-0.008} & 0.9197$\pm$0.0034 \gainpos{+0.0136} & 3.246$\pm$0.034 \gainpos{+0.064} & 2.153$\pm$0.015 \gainpos{+0.023} \\
\cmidrule(lr{1em}){2-8}

& RetinexMamba 
& 25.705$\pm$0.092 & 0.9345$\pm$0.0006 & 0.056$\pm$0.001 & 0.9123$\pm$0.0012 & 3.224$\pm$0.019 & 2.160$\pm$0.007 \\
& RetinexMamba with DITM 
& 25.920$\pm$0.157 \gainpos{+0.215} & 0.9417$\pm$0.0007 \gainpos{+0.0072} & 0.047$\pm$0.001 \gainpos{-0.009} & 0.9286$\pm$0.0016 \gainpos{+0.0163} & 3.329$\pm$0.017 \gainpos{+0.105} & 2.197$\pm$0.009 \gainpos{+0.037} \\
\cmidrule(lr{1em}){2-8}

& MambaLLIE 
& 25.229$\pm$0.090 & 0.9350$\pm$0.0018 & 0.054$\pm$0.002 & 0.9160$\pm$0.0061 & 3.243$\pm$0.038 & 2.186$\pm$0.009 \\
& MambaLLIE with DITM 
& 25.834$\pm$0.172 \gainpos{+0.605} & 0.9411$\pm$0.0013 \gainpos{+0.0061} & 0.047$\pm$0.002 \gainpos{-0.007} & 0.9280$\pm$0.0012 \gainpos{+0.0120} & 3.317$\pm$0.009 \gainpos{+0.074} & 2.203$\pm$0.003 \gainpos{+0.017} \\

\bottomrule
\end{tabular}}
\end{table*}

\subsection{Quantitative Evaluation}
\label{sec:quantitative_evaluation}

{\color{black}
\textbf{Performance on Paired Datasets.} Table~\ref{tab:paired_results} shows that DITM improves the mean performance in 124 of the 126 paired comparisons across three LOL datasets, seven LLIE backbones, and six evaluation metrics. Consistent gains are observed in the main full-reference metrics, including PSNR, SSIM, LPIPS, and DeepSSIM, while only minor degradations occur in IAA for MIRNet on LOLv1 and MambaLLIE on LOLv2-real. The three-run results further show that the mean improvements are consistently observed across most paired comparisons. The evaluated models cover CNN-, Transformer-, and Mamba-based architectures, demonstrating that DITM generalizes well across diverse supervised feed-forward LLIE backbones. The improvements obtained with CID-Net further indicate that DITM is complementary to existing denoising-oriented enhancement designs. 
}

\textbf{Performance on Unpaired Datasets.} 
We also conducted quantitative evaluation on five unpaired datasets. From the weighted average performance across the five datasets shown in Table~\ref{tab:unpaired_results}, all seven methods equipped with DITM improve the weighted average scores on the three metrics, except for the MUSIQ score of RetinexNet and CID-Net with DITM. These results suggest that DITM generally improves or maintains cross-dataset generalization on unpaired real-world low-light images, especially in terms of Q-Align IQA and IAA.

\begin{table*}[t]
\centering
\caption{Quantitative comparison on five unpaired datasets. The weighted average performance is obtained according to the image number of the five datasets.}
\label{tab:unpaired_results}
\small
\setlength{\tabcolsep}{3pt} 
\renewcommand{\arraystretch}{1.2}
\resizebox{\textwidth}{!}{
\begin{tabular}{l | ccc | ccc | ccc | ccc | ccc | ccc}
\toprule
\multicolumn{1}{c|}{\multirow{2}{*}{\textbf{Methods}}} & \multicolumn{3}{c|}{\textbf{DICM}} & \multicolumn{3}{c|}{\textbf{LIME}} & \multicolumn{3}{c|}{\textbf{MEF}} & \multicolumn{3}{c|}{\textbf{NPE}} & \multicolumn{3}{c|}{\textbf{VV}} & \multicolumn{3}{c}{\textbf{Weighted Average}} \\
\cmidrule(lr){2-4} \cmidrule(lr){5-7} \cmidrule(lr){8-10} \cmidrule(lr){11-13} \cmidrule(lr){14-16} \cmidrule(lr){17-19}
& MUSIQ$\uparrow$ & IQA$\uparrow$ & IAA$\uparrow$ & MUSIQ$\uparrow$ & IQA$\uparrow$ & IAA$\uparrow$ & MUSIQ$\uparrow$ & IQA$\uparrow$ & IAA$\uparrow$ & MUSIQ$\uparrow$ & IQA$\uparrow$ & IAA$\uparrow$ & MUSIQ$\uparrow$ & IQA$\uparrow$ & IAA$\uparrow$ & MUSIQ$\uparrow$ & IQA$\uparrow$ & IAA$\uparrow$ \\
\midrule

RetinexNet & 61.058 & 3.165 & 2.349 & 62.558 & 2.555 & 2.236 & 67.494 & 2.877 & 2.395 & 69.172 & 2.486 & 1.896 & 39.267 & 2.667 & 2.105 & 58.452 & 2.944 & 2.272 \\
RetinexNet with DITM & 61.112 & 3.225 & 2.368 & 61.729 & 2.628 & 2.269 & 67.111 & 2.952 & 2.459 & 68.542 & 2.617 & 2.007 & 37.892 & 2.779 & 2.158 & 58.067 \textcolor{green}{($\downarrow$)} & 3.021 \textcolor{red}{($\uparrow$)} & 2.310 \textcolor{red}{($\uparrow$)} \\
\cmidrule(lr){1-19}

MIRNet & 57.611 & 3.551 & 2.473 & 59.912 & 2.816 & 2.292 & 61.546 & 3.251 & 2.609 & 64.569 & 3.156 & 2.329 & 36.278 & 3.127 & 2.084 & 54.748 & 3.350 & 2.395 \\
MIRNet with DITM & 58.039 & 3.635 & 2.476 & 59.621 & 2.924 & 2.357 & 61.676 & 3.281 & 2.635 & 64.708 & 3.282 & 2.353 & 37.796 & 3.281 & 2.131 & 55.267 \textcolor{red}{($\uparrow$)} & 3.444 \textcolor{red}{($\uparrow$)} & 2.414 \textcolor{red}{($\uparrow$)} \\
\cmidrule(lr){1-19}

CID-Net & 58.324 & 3.619 & 2.399 & 59.359 & 2.939 & 2.397 & 62.508 & 3.463 & 2.598 & 66.436 & 3.475 & 2.436 & 40.264 & 3.606 & 2.230 & 56.081 & 3.534 & 2.396 \\
CID-Net with DITM & 58.165 & 3.659 & 2.439 & 59.672 & 3.022 & 2.407 & 62.377 & 3.459 & 2.626 & 66.986 & 3.516 & 2.469 & 39.925 & 3.625 & 2.241 & 55.973 \textcolor{green}{($\downarrow$)} & 3.567 \textcolor{red}{($\uparrow$)} & 2.426 \textcolor{red}{($\uparrow$)} \\

\midrule 

LLFormer & 58.983 & 3.743 & 2.527 & 57.597 & 3.320 & 2.450 & 63.144 & 3.485 & 2.603 & 65.904 & 3.598 & 2.593 & 32.546 & 3.568 & 2.161 & 54.903 & 3.634 & 2.466 \\
LLFormer with DITM & 59.316 & 3.775 & 2.539 & 58.187 & 3.275 & 2.450 & 63.032 & 3.497 & 2.623 & 66.480 & 3.616 & 2.560 & 32.983 & 3.579 & 2.175 & 55.232 \textcolor{red}{($\uparrow$)} & 3.652 \textcolor{red}{($\uparrow$)} & 2.476 \textcolor{red}{($\uparrow$)} \\
\cmidrule(lr){1-19}

RetinexFormer & 58.662 & 3.750 & 2.537 & 59.518 & 3.327 & 2.487 & 62.418 & 3.475 & 2.567 & 66.484 & 3.418 & 2.487 & 40.968 & 3.472 & 2.148 & 56.399 & 3.608 & 2.461 \\
RetinexFormer with DITM & 58.844 & 3.756 & 2.525 & 59.098 & 3.341 & 2.449 & 62.729 & 3.521 & 2.629 & 67.318 & 3.702 & 2.626 & 42.046 & 3.520 & 2.177 & 56.760 \textcolor{red}{($\uparrow$)} & 3.645 \textcolor{red}{($\uparrow$)} & 2.474 \textcolor{red}{($\uparrow$)} \\

\midrule

RetinexMamba & 58.894 & 3.780 & 2.548 & 60.734 & 3.325 & 2.466 & 61.832 & 3.405 & 2.530 & 66.212 & 3.486 & 2.507 & 59.349 & 3.583 & 2.200 & 59.971 & 3.639 & 2.469 \\
RetinexMamba with DITM & 58.883 & 3.806 & 2.558 & 60.575 & 3.285 & 2.442 & 62.684 & 3.525 & 2.649 & 66.845 & 3.496 & 2.523 & 59.979 & 3.683 & 2.233 & 60.223 \textcolor{red}{($\uparrow$)} & 3.686 \textcolor{red}{($\uparrow$)} & 2.498 \textcolor{red}{($\uparrow$)} \\
\cmidrule(lr){1-19}

MambaLLIE & 59.152 & 3.801 & 2.555 & 60.836 & 3.261 & 2.468 & 62.306 & 3.455 & 2.625 & 65.968 & 3.508 & 2.513 & 58.926 & 3.557 & 2.187 & 60.086 & 3.649 & 2.486 \\
MambaLLIE with DITM & 59.058 & 3.854 & 2.577 & 61.034 & 3.340 & 2.533 & 62.826 & 3.535 & 2.649 & 66.120 & 3.384 & 2.450 & 59.789 & 3.664 & 2.241 & 60.291 \textcolor{red}{($\uparrow$)} & 3.707 \textcolor{red}{($\uparrow$)} & 2.512 \textcolor{red}{($\uparrow$)} \\

\bottomrule
\end{tabular}
}
\end{table*}

\begin{table*}[t]
  \caption{Ablation study of alternative interpolation strategies using RetinexFormer as the baseline.}
  \label{tab:ablation_strategies}
  \centering
  \begin{tabular}{c|ccc|ccc|ccc}
    \toprule
    \multirow{2}{*}{Methods} & \multicolumn{3}{c|}{LOLv1} & \multicolumn{3}{c|}{LOLv2-real} & \multicolumn{3}{c}{LOLv2-synthetic} \\
    \cmidrule(r){2-4} \cmidrule(r){5-7} \cmidrule(r){8-10}
    & PSNR$\uparrow$ & SSIM$\uparrow$ & LPIPS$\downarrow$ & PSNR$\uparrow$ & SSIM$\uparrow$ & LPIPS$\downarrow$ & PSNR$\uparrow$ & SSIM$\uparrow$ & LPIPS$\downarrow$ \\
    \midrule
    Baseline & 23.498 & 0.8264 & 0.153 & 22.066 & 0.8378 & 0.178 & 25.149 & 0.9267 & 0.061 \\
    $B$-only      & 20.842 & 0.7966 & 0.165 & 20.370 & 0.8005 & 0.334 & 20.462 & 0.8766 & 0.125 \\
    ISW      & 23.381 & 0.8323 & 0.136 & 22.122 & 0.8447 & 0.154 & 25.497 & 0.9278 & 0.059 \\
    HDI      & 22.413 & 0.8333 & 0.127 & 21.903 & 0.8388 & 0.167 & 25.713 & 0.9346 & 0.053 \\
    \midrule
    \textbf{DITM (Ours)} & \textbf{24.050} & \textbf{0.8444} & \textbf{0.120} & \textbf{22.934} & \textbf{0.8532} & \textbf{0.139} & \textbf{25.717} & \textbf{0.9394} & \textbf{0.050} \\
    \bottomrule
  \end{tabular}
\end{table*}

\subsection{Visual Quality Evaluation}
\label{sec:visual_evaluation}

\begin{figure*}[htbp]
  \centering
  \includegraphics[width=\linewidth]{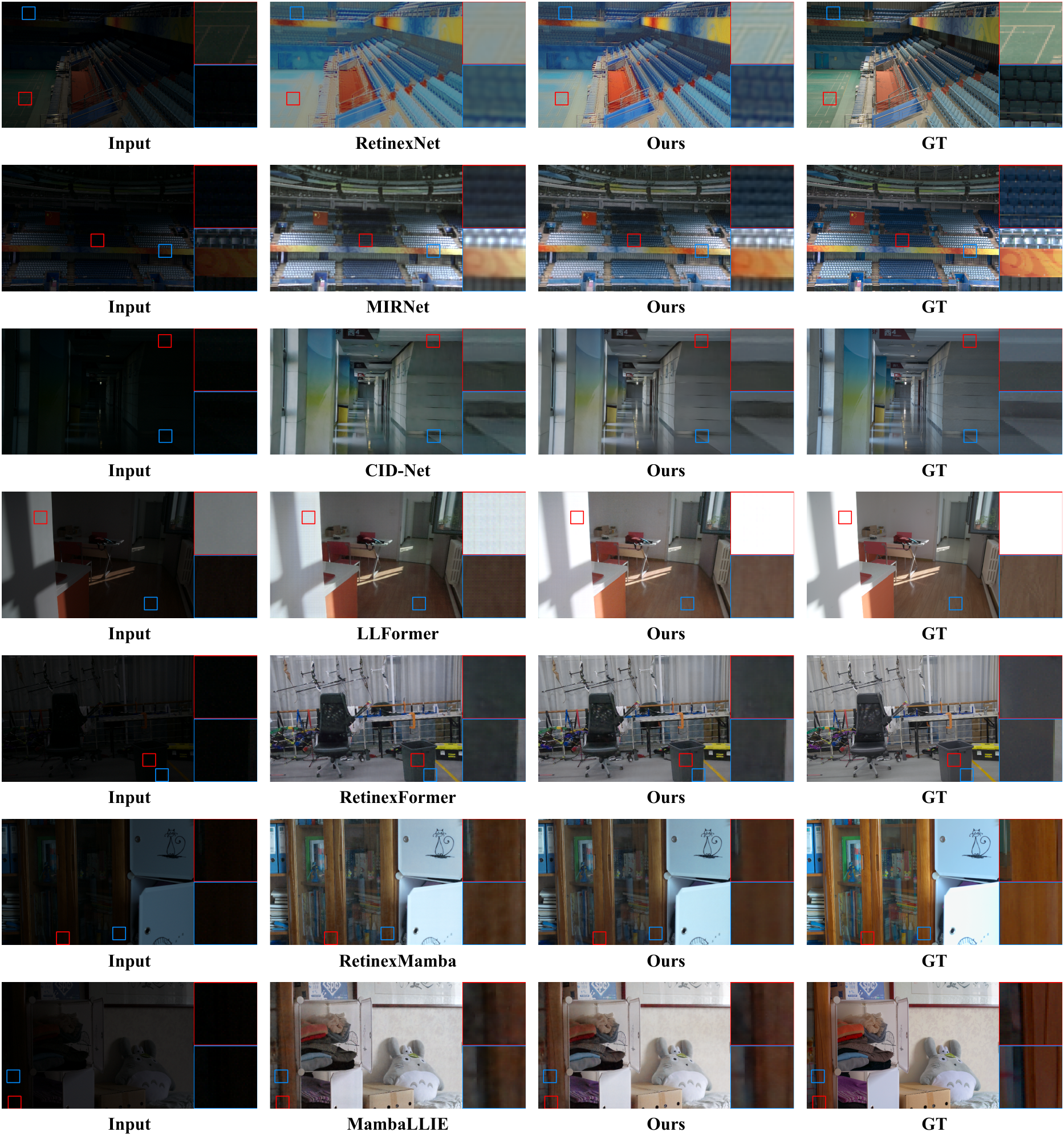}
  \caption{Visual comparison of different low-light image enhancement methods. From left to right: (a) Low-light Input, (b) Baseline model, (c) Ours (Baseline + DITM), and (d) Ground Truth. By integrating DITM into these existing supervised LLIE frameworks, the visual quality of the shown examples is improved in terms of artifact suppression, noise reduction, and preservation of structural details. Zoomed-in regions are provided for better visualization of local details.}
  \label{fig:paired_comparison}
\end{figure*}

\textbf{Visual Comparison on the Paired Datasets.} 
Fig.~\ref{fig:paired_comparison} provides visual comparisons on paired benchmarks. The baseline models improve visibility but still suffer from visible degradations in challenging low-light regions, including noise and structural artifacts. After integrating DITM, the enhanced results are closer to the ground truth, with reduced noise, fewer artifacts, and improved structural consistency. The zoomed-in regions in Fig.~\ref{fig:paired_comparison} provide a more detailed comparison. For CNN-based methods, DITM particularly aids in stabilizing color restoration while simultaneously suppressing chroma noise. In Transformer-based models, the improvement is particularly evident in reducing structural artifacts and recovering sharper edges. For Mamba-based methods, DITM further improves texture representation while reducing visual artifacts.

\begin{figure*}[htbp]
  \centering
  \includegraphics[width=\linewidth]{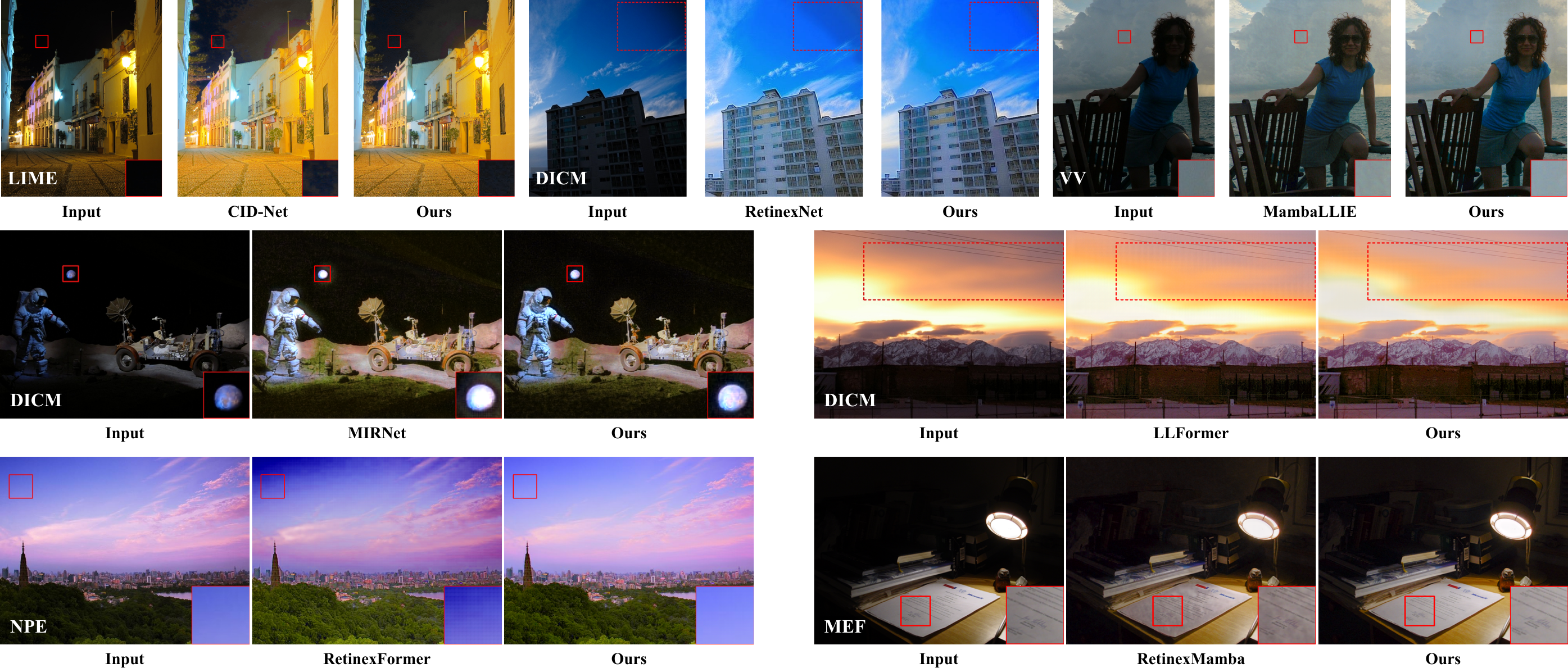}
  \caption{Visual comparison on unpaired real-world low-light datasets. From left to right: (a) Input, (b) Baseline, and (c) Ours (Baseline + DITM). The proposed DITM generally improves the generalization capability of existing supervised LLIE frameworks, resulting in improved visibility, reduced noise, and more robust restoration performance. Zoomed-in views are provided for a subset of regions for detailed comparison, while some regions are additionally highlighted using red dashed boxes to indicate noticeable artifacts.}
  \label{fig:unpaired_comparison}
\end{figure*}

\textbf{Visual Comparison on the Unpaired Datasets.}
As shown in Fig.~\ref{fig:unpaired_comparison}, DITM improves the robustness of different supervised LLIE models under real-world low-light conditions, producing more visually consistent results with fewer artifacts and better brightness balance. Baseline methods such as CID-Net introduce obvious halo artifacts around strongly illuminated regions in their enhanced results. For RetinexFormer, RetinexMamba and MIRNet, their results show varying degrees of local artifacts or exposure instability. In addition, LLFormer shows large-area stripe artifacts, while RetinexNet exhibits pronounced color blotches. These issues are alleviated after integrating DITM, as evidenced by the examples in Fig.~\ref{fig:unpaired_comparison}, indicating improved robustness.

\subsection{Further Analysis of DITM}
\label{sec:4:4}
\subsubsection{Comparison with Alternative Interpolation Strategies}
To verify that the gain comes from the proposed adaptive interpolation rather than from generic prediction interpolation, we compared DITM with several heuristic interpolation strategies on three paired datasets, using RetinexFormer as baseline. The interpolation strategies for comparison include:

\begin{itemize}
    \item \textbf{Baseline:} The original backbone network trained without interpolation.
    \item \textbf{Directly Using $B$ ($B$-only):} The model prediction $A$ is completely replaced by the mean brightness-aligned image $B$ during loss computation.
    \item \textbf{Interpolation with Static Weights (ISW):} A fixed fusion strategy with constant weights (0.8 for the model prediction $A$, 0.2 for the mean brightness-aligned image $B$) throughout training.
    \item \textbf{Heuristic Dynamic Interpolation (HDI):} A three-stage training schedule that first optimizes exclusively with $A$ (0–50k iterations), then linearly interpolates between the model prediction $A$ and the mean brightness-aligned image $B$ from 50k to 100k iterations using a linearly varying weight decreasing from 1 to 0, and finally transitions entirely to $B$ (100k–150k).
\end{itemize}

Experimental results based on different strategies are shown in Table~\ref{tab:ablation_strategies}. Across the three paired datasets (LOLv1, LOLv2-real, and LOLv2-synthetic), the interpolation strategy used in DITM consistently outperforms the others under different evaluation metrics, demonstrating the effectiveness of our dynamic image interpolation mechanism.

\begin{figure*}[htbp]
  \centering
  \includegraphics[width=\linewidth]{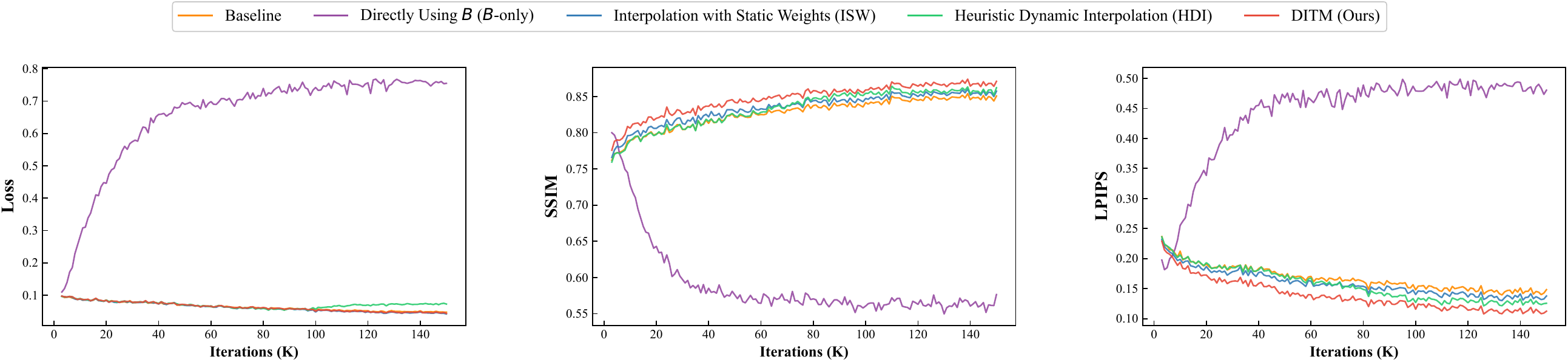}
  \caption{Comparison of training dynamics across different strategies on the LOLv1 training dataset. Curves of training loss, SSIM, and LPIPS are shown over 150K iterations, with values recorded every 1K iterations. Compared with the other strategies, DITM shows better performance over most of the training process.}

  \label{fig:trainingqvxian_comparison}
\end{figure*}

We further analyze how different strategies affect the optimization process. We conducted an experiment on the LOLv1 training dataset, and recorded the training loss, SSIM, and LPIPS over iterations during the training process, as shown in Fig.~\ref{fig:trainingqvxian_comparison}.  Among these strategies, $B$-only completely replaces the raw prediction $A$ with the brightness-aligned image $B$, thereby removing the direct supervision signal associated with the original brightness discrepancy of $A$. This weakens effective optimization and leads to degraded performance. Fig.~\ref{fig:trainingqvxian_comparison} shows that this strategy substantially degrades training stability and final performance. HDI adopts a manually designed interpolation strategy, whose heuristic scheduling limits adaptability across datasets. As ISW introduces a fixed interpolation weight, its static nature restricts flexibility under varying image conditions. In contrast, DITM determines the interpolation factor on a per-sample basis, where the weight $s$ is adaptively estimated for each image in the training set, leading to consistently better performance than other strategies.

\subsubsection{Hyperparameter Study}

\begin{figure*}[t]
    \centering
    \includegraphics[width=0.95\textwidth]{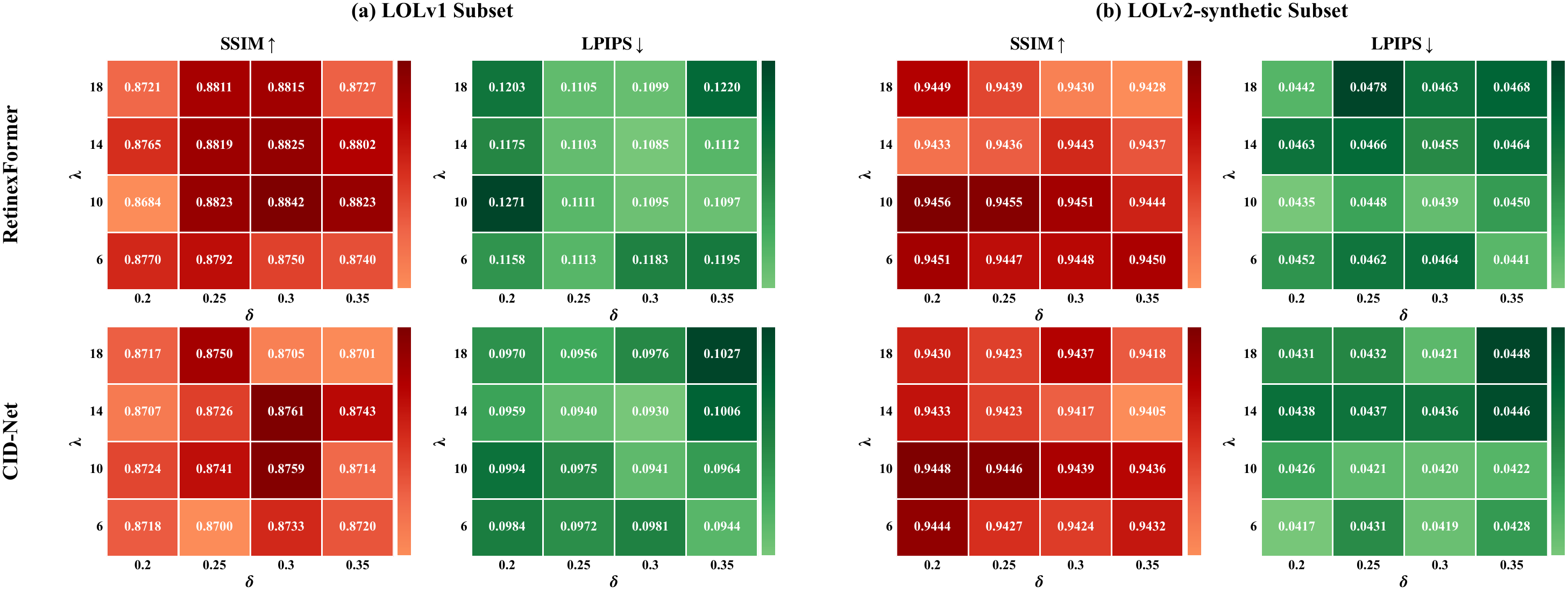}
    \caption{\textcolor{black}{SSIM and LPIPS of RetinexFormer and CID-Net under different $\lambda$ and $\delta$ settings on the held-out validation subsets of LOLv1 and LOLv2-synthetic.}}
    \label{fig:hyperparameter_analysis}
\end{figure*}

{\color{black}
To investigate the influence of hyperparameters $\lambda$ and $\delta$ on the overall performance, we conducted sensitivity experiments using RetinexFormer and CID-Net on validation subsets of LOLv1 and LOLv2-synthetic. For LOLv1, the original 485 training pairs were further split into 445 pairs for model training and 40 pairs for validating different hyperparameter settings. For LOLv2-synthetic, 800 of the original 900 training pairs were used for model training, and the remaining 100 pairs were used for validation. We evaluated $\lambda \in \{6, 10, 14, 18\}$ and $\delta \in \{0.2, 0.25, 0.3, 0.35\}$ and report the corresponding SSIM and LPIPS values in Fig.~\ref{fig:hyperparameter_analysis}.

The figure shows that model performance remains relatively stable across different hyperparameter combinations, with several settings achieving strong results, indicating that DITM is not overly sensitive to the choices of $\lambda$ and $\delta$. These results are also generally consistent with the roles of the two hyperparameters defined in Eq.~(\ref{eq:9}). Specifically, $\lambda$ controls the variation rate of $s$, while $\delta$ determines the transition region of the interpolation factor, around which $s$ changes more rapidly and the contribution of global brightness correction is gradually adjusted during training. Based on its consistently competitive validation performance across the four model–dataset settings, we select $(\lambda,\delta)=(10,0.3)$ as the unified configuration and use it in all other experiments without model- or dataset-specific retuning.

}

\begin{figure*}[!t]
  \centering
  \includegraphics[width=\linewidth]{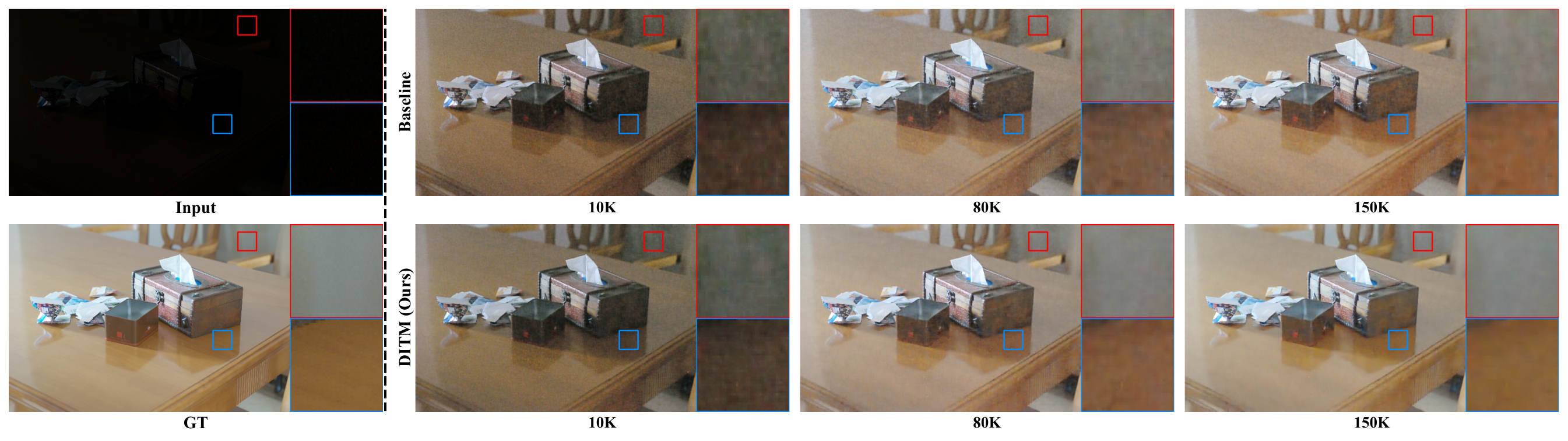}
  \caption{Qualitative comparison of the enhancement performance at different training stages (10K, 80K, and 150K iterations) on the LOLv1 training dataset. Compared to the baseline, the model with DITM begins to exhibit superior detail restoration and noise suppression from the middle stage of training. Zoomed-in views are provided to better illustrate the progressive improvements across different training stages.}
  \label{fig:training_comparison}

\end{figure*}

\begin{figure*}[!t]
  \centering
  \includegraphics[width=\linewidth]{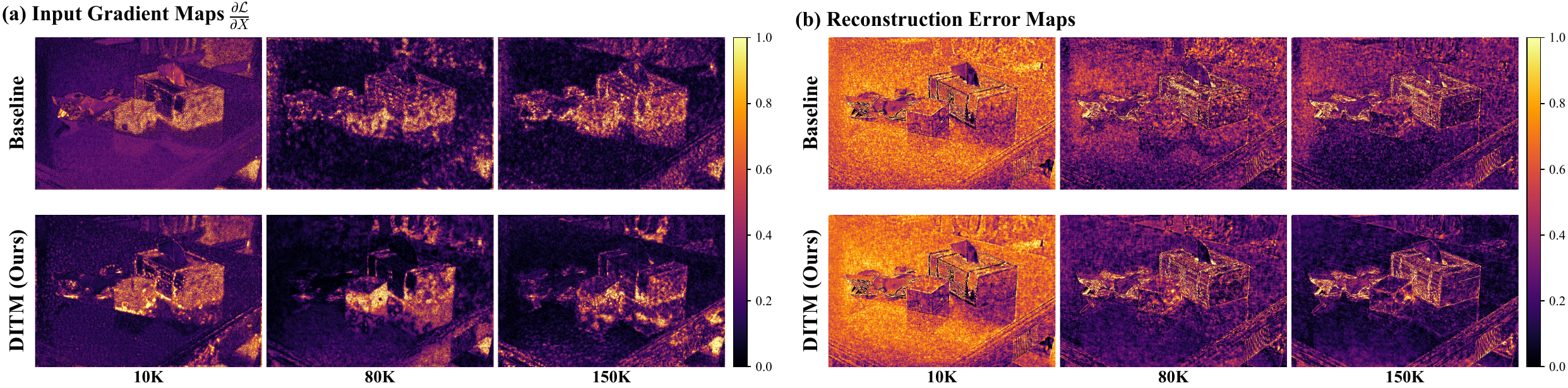}
  \caption{Visualization of input gradient maps $\frac{\partial \mathcal{L}}{\partial X}$ (gradients of the loss with respect to the input) and reconstruction error maps at different training stages (10K, 80K, and 150K iterations) for the Baseline (top row) and DITM (bottom row). (a) Input gradient maps suggest that DITM progressively yields more concentrated responses on structural regions, while the baseline exhibits more diffuse responses throughout training. (b) Reconstruction error maps illustrate that the errors gradually diminish during training for both methods. The model with DITM exhibits cleaner and more spatially uniform residual patterns in the later stage, suggesting a more focused training response in later stages.}
  \label{fig:vis_comparison}
\end{figure*}

\subsubsection{Analysis of Training Process with DITM}

We further investigated the training behavior associated with DITM. In particular, we aimed to provide empirical evidence consistent with the analysis in Section~\ref{sec:ditm}, and to examine how DITM affects model training across different stages. The experiment was conducted on the LOLv1 training dataset, using RetinexFormer as the baseline method. Fig.~\ref{fig:training_comparison} illustrates the enhancement results obtained from different training stages, while Fig.~\ref{fig:vis_comparison} presents input gradient and reconstruction error maps, jointly providing complementary insights into the training behavior associated with DITM.

At a typical early stage of training (10K iterations), the gradient maps of both models are widely distributed across the image (Fig.~\ref{fig:vis_comparison}(a)), with gradient responses appearing in both structural and background regions. This indicates that the network is still in an exploratory phase without a clearly localized training response. Correspondingly, the reconstruction error maps (Fig.~\ref{fig:vis_comparison}(b)) exhibit large errors across the entire image. As shown in Fig.~\ref{fig:training_comparison}, both models yield visibly under-enhanced results with less-suppressed noise.

At a typical middle stage of training (80K iterations), noticeable differences begin to emerge. As shown in Fig.~\ref{fig:vis_comparison}(a), the baseline continues to exhibit strong gradient responses around structural regions, while its responses in background areas remain scattered and relatively weak. In contrast, DITM produces more concentrated gradients on structurally relevant regions and suppresses irrelevant responses in the background. Correspondingly, the reconstruction error maps in Fig.~\ref{fig:vis_comparison}(b) show that the baseline still contains obvious residual errors, whereas DITM achieves more localized and reduced errors. These observations are further supported by the visual comparison in Fig.~\ref{fig:training_comparison}, where DITM exhibits clearer structures and less noise compared to the baseline. This behavior is consistent with the analysis in Section~\ref{sec:ditm}: as the adaptive factor $s$ evolves, the contribution of the mean brightness-aligned image $B$ branch increases, reducing the influence of global brightness discrepancies and making structural and texture-related residuals relatively more visible to the training objective.

At a typical late stage of training (150K iterations), the differences become more pronounced. The gradient responses of DITM are largely suppressed in non-structural regions (Fig.~\ref{fig:vis_comparison}(a)), indicating that the optimization has become more focused and stable. Meanwhile, the corresponding error maps (Fig.~\ref{fig:vis_comparison}(b)) are significantly cleaner and more uniform. In contrast, the baseline still shows residual errors and noticeable gradient responses in background areas, suggesting that the baseline still retains more residual optimization responses in background regions. As illustrated in Fig.~\ref{fig:training_comparison}, DITM produces more visually pleasing results with less noise and more consistent structural details.

Overall, these observations are consistent with the analysis in Section~\ref{sec:ditm}. The design of adaptive interpolation tries to reduce the influence of global brightness discrepancy as training progresses, and encourages the model to focus more on appearance-related details, such as structures and textures, at middle and late training stages.

{\color{black}

To quantitatively examine whether the adaptive interpolation factor $s$ evolves consistently with its intended design, we record its per-sample distribution during the training of RetinexFormer at 10K, 80K, and 150K iterations on LOLv1, LOLv2-real, and LOLv2-synthetic. In addition to the violin distributions, we report the mean value $\mu_s$ and the low-factor ratio.

\begin{equation}
R_{s<0.1}
=
\frac{1}{N}
\sum_{i=1}^{N}
\mathbb{I}(s_i<0.1)
\times 100\%,
\end{equation}
where $N$ denotes the number of recorded samples and $\mathbb{I}(\cdot)$ is the indicator function. A larger $R_{s<0.1}$ indicates that a greater proportion of samples assign a stronger contribution to the mean brightness-aligned prediction $B$.

\begin{figure*}[t]
    \centering
    \includegraphics[width=0.92\textwidth]{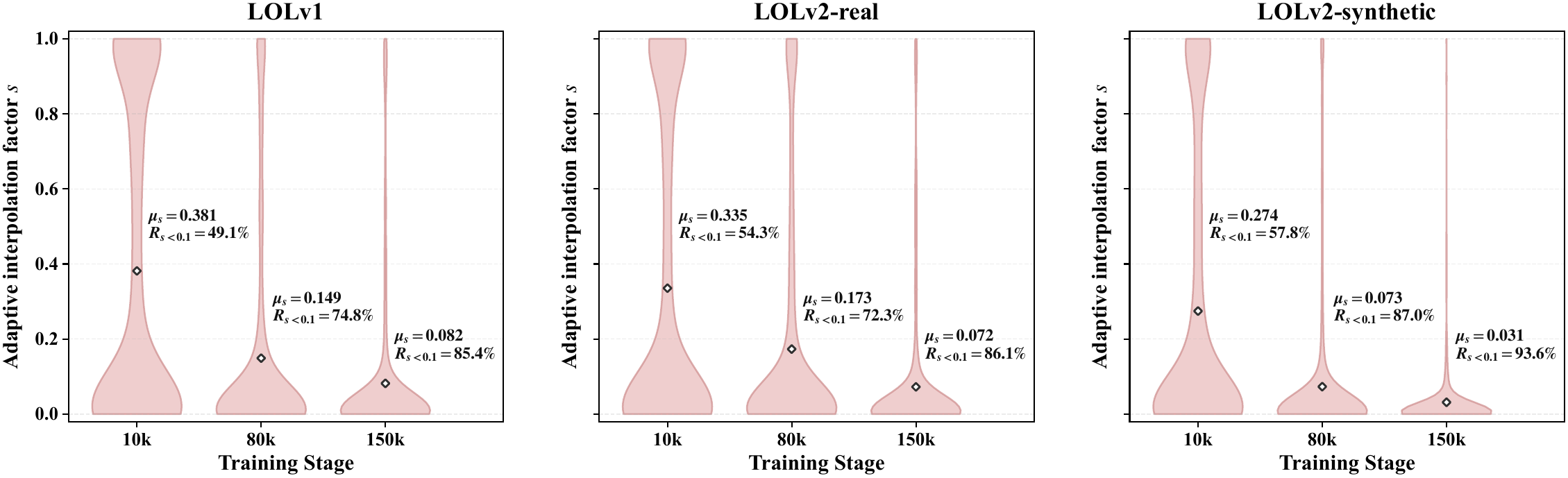}
    \caption{\textcolor{black}{
    Distribution evolution of the adaptive interpolation factor $s$ during the training of RetinexFormer on LOLv1, LOLv2-real, and LOLv2-synthetic. The distributions are recorded at 10K, 80K, and 150K training iterations. The width of each violin represents the distribution density of the per-sample interpolation factors, while the diamond marker denotes the corresponding mean value $\mu_s$. We additionally report $R_{s<0.1}$, which denotes the proportion of recorded samples whose interpolation factors are smaller than 0.1 at each training stage. Across all three datasets, the dominant distribution of $s$ progressively shifts toward smaller values, accompanied by decreasing $\mu_s$ and increasing $R_{s<0.1}$. The remaining thin upper tails indicate that a small number of samples retain relatively large $s$ values at later stages, reflecting the per-sample adaptive behavior of DITM rather than a fixed iteration-dependent schedule.
    }}
    \label{fig:s_distribution}
\end{figure*}

As shown in Fig.~\ref{fig:s_distribution}, the distribution mass of $s$ progressively shifts toward smaller values on all three datasets. On LOLv1, $\mu_s$ decreases from 0.381 at 10K iterations to 0.149 at 80K and 0.082 at 150K, while $R_{s<0.1}$ increases from 49.1\% to 74.8\% and 85.4\%. On LOLv2-real, $\mu_s$ decreases from 0.335 to 0.173 and 0.072, while $R_{s<0.1}$ increases from 54.3\% to 72.3\% and 86.1\%. On LOLv2-synthetic, $\mu_s$ decreases from 0.274 to 0.073 and 0.031, while $R_{s<0.1}$ increases from 57.8\% to 87.0\% and 93.6\%.

These statistics verify the intended stage-dependent behavior of DITM. At the earlier training stage, relatively larger values of $s$ keep the interpolated prediction $C$ closer to the raw prediction $A$, preserving stronger direct supervision for correcting substantial brightness discrepancy. As training progresses, the dominant distribution of $s$ shifts toward smaller values, moving $C$ closer to the mean brightness-aligned prediction $B$. This reduces the relative influence of residual image-level luminance mismatch in the loss-side prediction and makes non-global residual errors, including noise, texture degradation, structural artifacts, and local exposure mismatch, relatively more involved in subsequent optimization.

The thin upper tails observed at later stages indicate that a small number of samples still retain relatively large $s$ values. This behavior is consistent with the per-sample design of DITM: samples with larger remaining brightness discrepancies preserve a stronger contribution from the raw prediction $A$, rather than following a fixed iteration-dependent schedule. Together with the gradient-response analysis, these statistics provide complementary evidence that DITM progressively reshapes the composition of the supervised error signal as intended.
}

\subsubsection{\textcolor{black}{Robustness under Extremely Low-Light Conditions}}

\begin{table}[t]
    \centering
    \caption{\textcolor{black}{Numerical behavior of DITM on the darkest 10\% samples of the LOLv1 training set at different training stages.}}
    \label{tab:darkest10_robustness}
    \renewcommand{\arraystretch}{1.08}
    \setlength{\tabcolsep}{3.5pt}
    \resizebox{\columnwidth}{!}{%
        \begin{tabular}{ccccccc}
            \toprule
            \textbf{Model} & \textbf{Stage}
            & $\boldsymbol{\min(\mu_A)}$
            & $\boldsymbol{\operatorname{Mean}(\mu_A)}$
            & $\boldsymbol{\operatorname{Mean}(\beta)}$
            & $\boldsymbol{\max(\beta)}$
            & $\boldsymbol{\operatorname{Mean}(s)}$ \\
            \midrule
            \multirow{3}{*}{RetinexFormer with DITM}
            & Early  & 0.1794 & 0.3634 & 1.1964 & 1.7028 & 0.5739 \\
            & Middle & 0.1896 & 0.4044 & 1.0787 & 1.4301 & 0.2617 \\
            & Late   & 0.1990 & 0.4128 & 1.0510 & 1.2256 & 0.1163 \\
            \midrule
            \multirow{3}{*}{RetinexMamba with DITM}
            & Early  & 0.1694 & 0.3542 & 1.2250 & 1.6734 & 0.6324 \\
            & Middle & 0.1902 & 0.3923 & 1.1027 & 1.4034 & 0.3161 \\
            & Late   & 0.2064 & 0.4103 & 1.0603 & 1.3587 & 0.1354 \\
            \bottomrule
        \end{tabular}%
    }
\end{table}

\textcolor{black}{To further examine the numerical behavior of DITM under challenging low-light conditions, we selected the darkest 10\% of the LOLv1 training set according to the mean brightness of the low-light inputs, resulting in 49 samples. The same fixed subset was evaluated using early-, middle-, and late-stage checkpoints obtained from RetinexFormer and RetinexMamba trained with DITM. As shown in Table~\ref{tab:darkest10_robustness}, $\mu_A$ remained numerically well behaved across all settings. The minimum observed $\mu_A$ ranged from $0.1694$ to $0.2064$, while the maximum observed $\beta=\frac{\mu_Y}{\mu_A}$ remained below $1.7028$. Moreover, the mean $\beta$ progressively decreased from $1.1964$ to $1.0510$ for RetinexFormer and from $1.2250$ to $1.0603$ for RetinexMamba as training proceeded. No NaN or Inf values were observed. These results indicate that the luminance statistics and the resulting scaling ratio remain numerically stable within the evaluated challenging low-light subset. Meanwhile, the mean value of $s$ decreased from $0.5739$ to $0.1163$ for RetinexFormer and from $0.6324$ to $0.1354$ for RetinexMamba, consistent with the intended transition from preserving the raw prediction at the early stage to increasingly incorporating the brightness-aligned prediction as the brightness discrepancy is reduced.
}

\subsubsection{\textcolor{black}{Performance under Spatially Non-Uniform Illumination}}
\textcolor{black}{
We further evaluate DITM on samples with pronounced spatially non-uniform illumination. To construct this evaluation subset, each low-light input $X$ is first converted into a grayscale brightness map and then smoothed to capture its low-frequency illumination distribution. The illumination non-uniformity score is defined as}
\begin{equation}
{\color{black}
U(X)=
\frac{\operatorname{Std}(M(X))}
{\operatorname{Mean}(M(X))},
}
\end{equation}\textcolor{black}{where $M(X)$ denotes the Gaussian-smoothed brightness map of $X$ obtained with a $51\times51$ Gaussian kernel. A larger $U(X)$ indicates stronger relative spatial variation in the low-frequency brightness distribution. We rank the LOLv2-synthetic test images according to $U(X)$ and select the top 20\% as the spatially non-uniform subset. This score is used only for subset construction rather than as an image-quality metric.}

\begin{table*}[t]
\caption{\textcolor{black}{Performance comparison on the top 20\% spatially non-uniform samples from the LOLv2-synthetic test set. Each setting was independently repeated three times, and the mean and standard deviation are reported. Values in parentheses denote the mean change of DITM over the corresponding baseline. Red indicates improvement, while green indicates degradation.}}
\label{tab:nonuniform_results}
\centering
\scriptsize
\setlength{\tabcolsep}{3.2pt}
\resizebox{\textwidth}{!}{
\begin{tabular}{>{\raggedright\arraybackslash}p{3.8cm}|cccccc}
\toprule
\multicolumn{1}{c|}{\textbf{Methods}}
& \multicolumn{6}{c}{\textbf{Evaluation Metrics}} \\
& PSNR$\uparrow$
& SSIM$\uparrow$
& LPIPS$\downarrow$
& DeepSSIM$\uparrow$
& IQA$\uparrow$
& IAA$\uparrow$ \\
\midrule

RetinexFormer
& 27.012$\pm$0.607
& 0.9035$\pm$0.0025
& 0.073$\pm$0.002
& 0.8823$\pm$0.0005
& 3.014$\pm$0.009
& 2.176$\pm$0.004 \\

RetinexFormer with DITM
& 27.673$\pm$0.462 \gainpos{+0.661}
& 0.9121$\pm$0.0008 \gainpos{+0.0086}
& 0.068$\pm$0.002 \gainpos{-0.005}
& 0.8928$\pm$0.0015 \gainpos{+0.0105}
& 3.070$\pm$0.040 \gainpos{+0.056}
& 2.205$\pm$0.015 \gainpos{+0.029} \\

\cmidrule(lr{1em}){1-7}

RetinexMamba
& 26.264$\pm$0.577
& 0.9036$\pm$0.0003
& 0.072$\pm$0.001
& 0.8845$\pm$0.0024
& 3.108$\pm$0.016
& 2.256$\pm$0.009 \\

RetinexMamba with DITM
& 26.596$\pm$0.051 \gainpos{+0.332}
& 0.9124$\pm$0.0003 \gainpos{+0.0088}
& 0.065$\pm$0.001 \gainpos{-0.007}
& 0.8956$\pm$0.0018 \gainpos{+0.0111}
& 3.231$\pm$0.022 \gainpos{+0.123}
& 2.311$\pm$0.011 \gainpos{+0.055} \\

\bottomrule
\end{tabular}}
\end{table*}

\begin{table*}[!t]
    \caption{Evaluation of DITM on diffusion-based low-light image enhancement methods on the LOLv1 dataset.}
    \label{tab:diffusionmodel_lolv1_results}
    \centering
    \footnotesize
    \setlength{\tabcolsep}{4pt}
    \renewcommand{\arraystretch}{1.12}
    \resizebox{0.95\textwidth}{!}{
    \begin{tabular}{l|cccccc}
    \toprule
    \multicolumn{1}{c|}{Methods}
    & PSNR$\uparrow$
    & SSIM$\uparrow$
    & DeepSSIM$\uparrow$
    & LPIPS$\downarrow$
    & IQA$\uparrow$
    & IAA$\uparrow$ \\
    \midrule
    
    PyDiff
    & 21.958
    & 0.8969
    & 0.8283
    & 0.110
    & 4.158
    & 2.245 \\
    
    PyDiff with DITM
    & 22.004 \textcolor{red}{(+0.046)}
    & 0.8976 \textcolor{red}{(+0.0007)}
    & 0.8289 \textcolor{red}{(+0.0006)}
    & 0.110 \textcolor{black}{(+0.000)}
    & 4.160 \textcolor{red}{(+0.002)}
    & 2.241 \textcolor{green}{(-0.004)} \\
    
    \cmidrule(lr{1em}){1-7}
    
    Diff-Retinex++
    & 24.344
    & 0.8655
    & 0.8308
    & 0.105
    & 4.201
    & 2.282 \\
    
    Diff-Retinex++ with DITM
    & 24.108 \textcolor{green}{(-0.236)}
    & 0.8672 \textcolor{red}{(+0.0017)}
    & 0.8326 \textcolor{red}{(+0.0018)}
    & 0.103 \textcolor{red}{(-0.002)}
    & 4.213 \textcolor{red}{(+0.012)}
    & 2.281 \textcolor{green}{(-0.001)} \\
    
    \bottomrule
    \end{tabular}
    }
\end{table*}

\textcolor{black}{As shown in Table~\ref{tab:nonuniform_results}, the selected spatially non-uniform subset exhibits different performance characteristics compared with the full LOLv2-synthetic test set, with relatively lower SSIM and higher LPIPS values, indicating the influence of spatial illumination variation on image restoration. Nevertheless, DITM consistently improves both RetinexFormer and RetinexMamba on this subset. Across three independent runs, DITM achieves consistent improvements in SSIM, LPIPS, DeepSSIM, IQA, and IAA, while also providing positive average PSNR gains. These results suggest that although DITM is guided by global luminance statistics and does not explicitly model spatial illumination distributions, the potential limitation does not significantly affect its effectiveness under the evaluated spatially non-uniform illumination conditions.}

\begin{table}[t]
\centering
\caption{Training-time overhead of DITM on the adopted LLIE backbones in our experiments.}
\label{tab:training_overhead}
\scriptsize
\setlength{\tabcolsep}{3pt}
\renewcommand{\arraystretch}{1.12}
\resizebox{\columnwidth}{!}{
\begin{tabular}{lcc}
\toprule
\textbf{Model} & \textbf{Training Time / Iter} & \textbf{Peak GPU Memory} \\
\midrule
RetinexNet & 326.458 ms & 2.103 GB \\
RetinexNet with DITM & 328.990 ms {\scriptsize (+0.78\%)} & 2.104 GB \\
\midrule
MIRNet & 519.327 ms & 20.040 GB \\
MIRNet with DITM & 520.944 ms {\scriptsize (+0.31\%)} & 20.041 GB \\
\midrule
CID-Net & 210.626 ms & 14.350 GB \\
CID-Net with DITM & 213.715 ms {\scriptsize (+1.47\%)} & 14.375 GB \\
\midrule
LLFormer & 425.855 ms & 20.060 GB \\
LLFormer with DITM & 427.666 ms {\scriptsize (+0.43\%)} & 20.063 GB \\
\midrule
RetinexFormer & 48.034 ms & 3.008 GB \\
RetinexFormer with DITM & 49.301 ms {\scriptsize (+2.64\%)} & 3.008 GB \\
\midrule
MambaLLIE & 134.341 ms & 4.638 GB \\
MambaLLIE with DITM & 135.370 ms {\scriptsize (+0.77\%)} & 4.641 GB \\
\midrule
RetinexMamba & 266.436 ms & 9.871 GB \\
RetinexMamba with DITM & 268.306 ms {\scriptsize (+0.70\%)} & 9.873 GB \\
\bottomrule
\end{tabular}
}
\end{table}

\subsubsection{Training-Time Overhead}
We further evaluate the efficiency of DITM on seven LLIE backbones, covering CNN-, Transformer-, and Mamba-based architectures. As reported in Table~\ref{tab:training_overhead}, DITM only brings marginal training-time overhead, with the largest increase being 2.64\%. The peak GPU memory remains almost unchanged. Since DITM is removed during inference, the deployed models have the same parameters, FLOPs, and latency as the original backbones.

{\color{black}
\subsection{Applicability Scope and Limitations of DITM}
\label{sec:applicability}

DITM is designed for paired supervised LLIE models that produce an explicit image-space prediction and directly optimize it against a corresponding normal-light reference. Within this setting, DITM is independent of the specific backbone architecture, as demonstrated by its consistent effectiveness across CNN-, Transformer-, and Mamba-based feed-forward models. Its applicability, however, depends on both the availability of paired supervision and the role of image-space reconstruction in the underlying optimization formulation.

\textbf{Requirement for Paired Supervision.}
Both the mean brightness-aligned prediction $B$ and the adaptive interpolation factor $s$ require the corresponding normal-light ground-truth image $Y$. Therefore, DITM in its current formulation is not directly applicable to unpaired, unsupervised, or zero-reference LLIE training, where such a reference is unavailable during optimization. This limitation concerns the training regime rather than the testing data. A model trained with DITM under paired supervision can still be directly evaluated on unpaired real-world test sets, as demonstrated by our cross-dataset evaluation.

\begin{table}[t]
\centering
\caption{
\textcolor{black}{Gradient response analysis of DITM on diffusion-based LLIE models.
The results are averaged over sampled images.
$\rho_g$ and $\cos$ denote the gradient norm ratio and cosine similarity between baseline and DITM optimization, respectively.}
}
\label{tab:diffusion_gradient}

\renewcommand{\arraystretch}{1.08}
\setlength{\tabcolsep}{2pt}

\resizebox{\columnwidth}{!}{
\begin{tabular}{c|l|cc|cc}
\toprule

\multirow{2}{*}{Method}
&
\multicolumn{1}{c|}{\multirow{2}{*}{Parameter Group}}
&
\multicolumn{2}{c|}{Early Stage}
&
\multicolumn{2}{c}{Late Stage}
\\

&
&
$\rho_g$
&
$\cos$
&
$\rho_g$
&
$\cos$
\\

\midrule

\multirow{3}{*}{PyDiff}
&
Full Model
&
0.870
&
0.543
&
0.862
&
0.616
\\

&
Diffusion Denoising Module
&
1.000
&
1.000
&
1.000
&
1.000
\\

&
Image Restoration Module
&
0.860
&
0.516
&
0.841
&
0.585
\\

\midrule

\multirow{3}{*}{Diff-Retinex++}
&
Full Model
&
0.987
&
0.981
&
0.978
&
0.976
\\

&
Diffusion Denoising Module
&
1.000
&
1.000
&
1.000
&
1.000
\\

&
Image Restoration Module
&
0.976
&
0.962
&
0.974
&
0.971
\\

\bottomrule

\end{tabular}
}
\end{table}

\textbf{Dependence on the Restoration Formulation.}
DITM is most effective when the reshaped image prediction directly participates in optimizing the parameters responsible for the final restoration. To examine this condition, we further evaluated DITM on two diffusion-based LLIE methods, PyDiff~\cite{zhao2023pydiff} and Diff-Retinex++~\cite{yi2025diff}, on the LOLv1 dataset. For each method, DITM was incorporated into the image-space reconstruction branch as a training-only prediction reshaping operation, while all other loss terms, loss weights, and training settings were kept unchanged. The quantitative results are reported in Table~\ref{tab:diffusionmodel_lolv1_results}.

As shown in Table~\ref{tab:diffusionmodel_lolv1_results}, unlike the consistent improvements observed on supervised feed-forward LLIE backbones, DITM provides only marginal and inconsistent gains on the two investigated diffusion-based models. To further analyze this phenomenon, we compare the gradient responses induced by the original and DITM-modified objectives under the same model state. Specifically, for a parameter subset $\theta_m$, we compute the gradient norm ratio

\begin{equation}
\rho_g(\theta_m)
=
\frac{
\|\nabla_{\theta_m}\mathcal{L}_{\mathrm{DITM}}\|_2
}{
\|\nabla_{\theta_m}\mathcal{L}_{\mathrm{base}}\|_2
},
\end{equation}
and the cosine similarity

\begin{equation}
\cos(\theta_m)
=
\frac{
\nabla_{\theta_m}\mathcal{L}_{\mathrm{base}}
\cdot
\nabla_{\theta_m}\mathcal{L}_{\mathrm{DITM}}
}{
\|\nabla_{\theta_m}\mathcal{L}_{\mathrm{base}}\|_2
\|\nabla_{\theta_m}\mathcal{L}_{\mathrm{DITM}}\|_2
},
\end{equation} where $\mathcal{L}_{\mathrm{base}}$ and $\mathcal{L}_{\mathrm{DITM}}$ denote the original and DITM-modified objectives, respectively. Here, $\theta_m$ denotes a specific parameter group, including the full model, diffusion denoising module, or image restoration module. The results are reported in Table~\ref{tab:diffusion_gradient}.

These gradient observations provide a quantitative explanation for the limited improvements observed on the two diffusion-based LLIE models. Unlike supervised feed-forward models, where the reshaped prediction directly supervises the backbone responsible for producing the restored image, the investigated diffusion models employ separate optimization pathways for diffusion denoising and image-space reconstruction. As shown in Table~\ref{tab:diffusion_gradient}, the gradients of the diffusion denoising modules remain nearly unchanged after introducing DITM, indicating that the reshaped reconstruction supervision does not establish an effective optimization connection with the parameters responsible for the diffusion denoising process. Therefore, the effectiveness of DITM depends on whether the reshaped reconstruction supervision can influence the parameters governing the final restoration process.

These observations indicate that DITM should be regarded as a backbone-agnostic prediction reshaping strategy within paired supervised feed-forward LLIE, rather than a universally applicable mechanism for all LLIE paradigms. Diffusion models without explicit image-space reconstruction objectives~\cite{hou2023global,shang2024multi} cannot directly adopt the current formulation. For diffusion-based models containing image-space reconstruction branches~\cite{zhao2023pydiff,yi2025diff}, the benefit of DITM depends on the optimization connection between the reshaped supervision and the diffusion restoration process.
}

\section{Conclusion}

We presented a lightweight prediction-reshaping strategy called DITM for supervised feed-forward LLIE. Without modifying the backbone architecture or introducing additional inference-time cost, DITM dynamically interpolates the raw model prediction with a brightness-aligned prediction during training. This operation reduces the relative dominance of global luminance discrepancy in the supervised prediction, allowing residual artifact-, structure-, and detail-related errors to become more involved in subsequent optimization. Extensive experiments on multiple benchmark datasets demonstrate that DITM generally provides stable quantitative and qualitative improvements across CNN-, Transformer-, and Mamba-based LLIE backbones. With its lightweight training-time design and zero-overhead inference property, DITM offers a simple way to improve supervised feed-forward LLIE models.

DITM is currently designed for paired supervised LLIE, as it relies on the ground-truth mean brightness during training. Future work will explore reliable pseudo brightness targets or learned brightness references to extend the prediction-reshaping strategy to unpaired, self-supervised, and zero-reference LLIE. Since the current global mean-brightness alignment does not explicitly model spatially varying illumination, future work will explore spatially adaptive illumination guidance to better handle non-uniform illumination and local lighting variations. In addition, it is important to investigate prediction reshaping in diffusion-based restoration frameworks where image-space reconstruction supervision remains gradient-connected to the denoising network, particularly its appropriate placement along the denoising trajectory.

\vspace{0.2ex}{
\noindent{\textbf{Acknowledgments.} This work was supported by the National Natural Science Foundation of China under Grant XXXXXXX.}}


\bibliographystyle{IEEEtran}
\bibliography{DITMbib}



\vspace{-1.5em}
\begin{IEEEbiography}[{\includegraphics[width=1in,height=1.25in,clip,keepaspectratio]{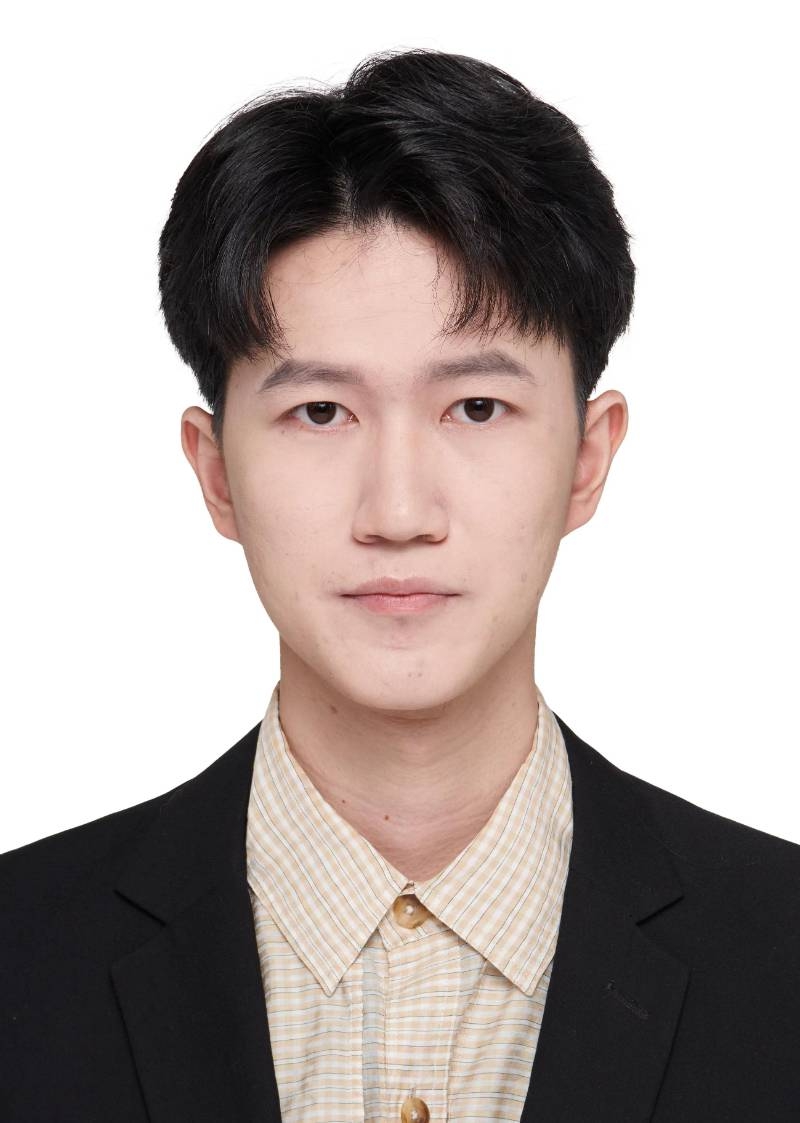}}]{Peirui~Zhu}
received the B.S. degree from Hefei University of Technology (HFUT) in 2024, where he is currently pursuing the M.S. degree with the School of Computer Science and Information Engineering. His research interests include low-light image enhancement.
\end{IEEEbiography}
\vspace{-3em}

\begin{IEEEbiography}[{\includegraphics[width=1in,height=1.25in,clip,keepaspectratio]{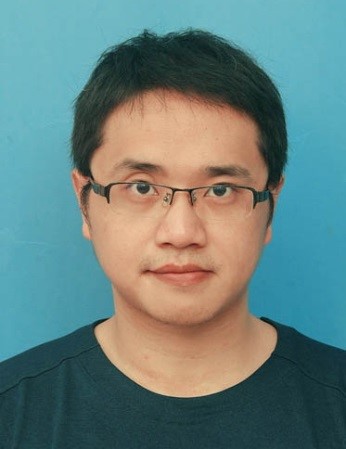}}]{Shijie~Hao}
is a Professor with the School of Computer Science and Information Engineering, Hefei University of Technology (HFUT). He received the Ph.D. degree from HFUT in 2012. He was a visiting scholar with the University of North Carolina at Chapel Hill (UNC) from 2015 to 2016. His research interests include low-level image processing.
\end{IEEEbiography}
\vspace{-3em}

\begin{IEEEbiography}[{\includegraphics[width=1in,height=1.25in,clip,keepaspectratio]{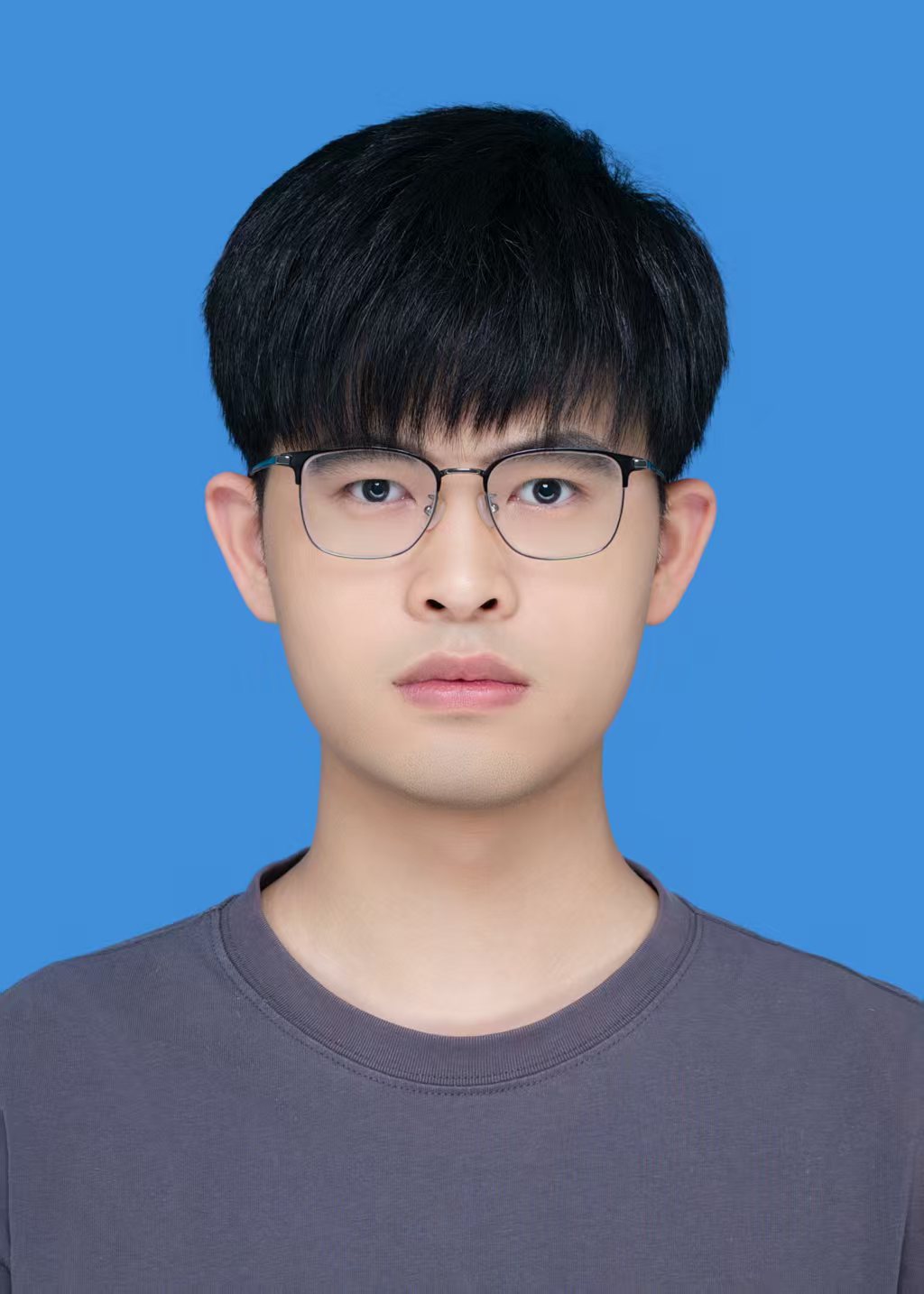}}]{Jingxi~Liao}
received the B.S. and M.S. degrees from the School of Computer Science and Information Engineering, Hefei University of Technology (HFUT). His research interests include low-light image enhancement.
\end{IEEEbiography}
\vspace{-3em}

\begin{IEEEbiography}[{\includegraphics[width=1in,height=1.25in,clip,keepaspectratio]{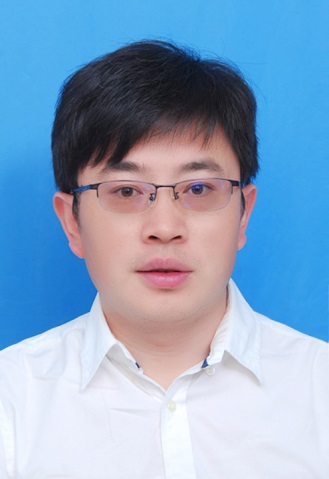}}]{Richang~Hong}
received the Ph.D. degree from the University of Science and Technology of China (USTC) in 2008. From 2008 to 2010, he was a Research Fellow with the National University of Singapore. He is currently a Professor with HFUT. He has authored over 200 publications in multimedia content analysis and related areas. He serves as an Associate Editor for IEEE TCSVT, IEEE TBD, IEEE TMM, and ACM TOMM.
\end{IEEEbiography}
\vspace{-3em}

\begin{IEEEbiography}[{\includegraphics[width=1in,height=1.25in,clip,keepaspectratio]{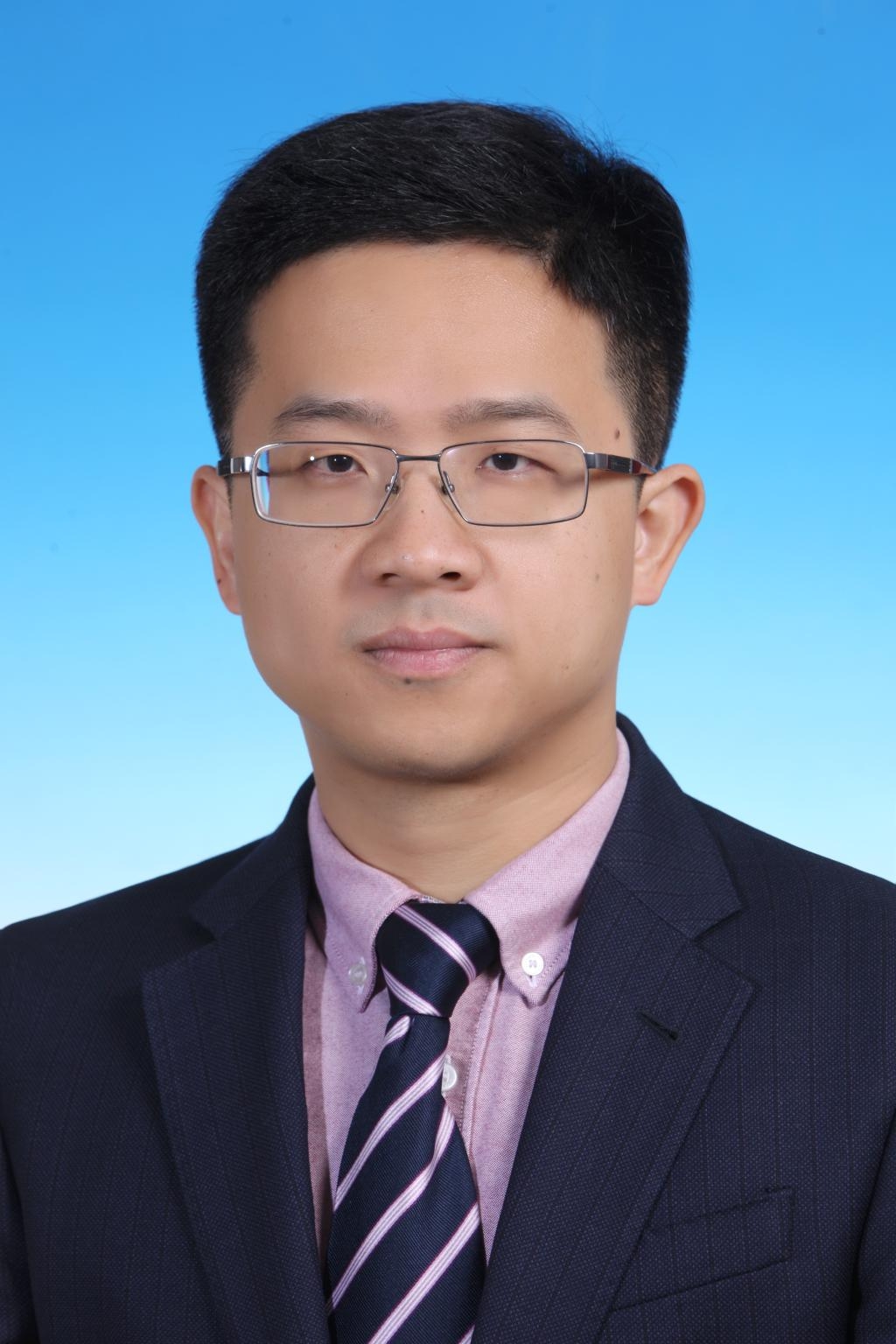}}]{Meng~Wang}
received the Ph.D. degree from the University of Science and Technology of China (USTC) in 2008. He is currently a Professor with HFUT. His research interests include multimedia, computer vision, and pattern recognition. He has authored over 200 papers in these areas. He is a Fellow of the IEEE and serves/served as an AE for IEEE TPAMI, IEEE TCSVT, IEEE TKDE, IEEE TNNLS, and IEEE TMM.
\end{IEEEbiography}


\end{document}